\pdfoutput=1

\documentclass[11pt,dvipsnames]{article}

\usepackage[final]{eacl/eacl2023}

\usepackage{times}
\usepackage{latexsym}

\usepackage[T1]{fontenc}

\usepackage[utf8]{inputenc}

\usepackage{microtype}

\usepackage{inconsolata}

\usepackage{url}            
\usepackage{booktabs}       
\usepackage{amsfonts}       
\usepackage{nicefrac}       
\usepackage{xcolor} 

\definecolor{myGray}{rgb}{0.97,0.97,0.97}

\usepackage{wrapfig}

\usepackage{ownstyles}

\usepackage{booktabs}
\usepackage{nicematrix}
\usepackage{mathtools}
\usepackage{multirow}
\usepackage{makecell}
\usepackage{array}
\usepackage{subcaption}
\usepackage{tabularx,colortbl}
\usepackage{enumitem}
\usepackage{commath}
\usepackage{amsfonts}
\usepackage{soul}
\usepackage{tikz}

\usepackage{amsmath,amsfonts,bm}

\def\eqref#1{equation~\ref{#1}}

\def\1{\bm{1}}

\def\vd{{\bm{d}}}

\def\vp{{\bm{p}}}
\def\vq{{\bm{q}}}

\DeclareMathAlphabet{\mathsfit}{\encodingdefault}{\sfdefault}{m}{sl}
\SetMathAlphabet{\mathsfit}{bold}{\encodingdefault}{\sfdefault}{bx}{n}

\definecolor{magenta(dye)}{rgb}{0.79, 0.08, 0.48}
\definecolor{mycitecolor}{rgb}{0,0.08,0.75} 

\usepackage{float}
\usepackage{enumitem}
\usepackage{listings}

\definecolor{codegreen}{rgb}{0,0.6,0}
\definecolor{codegray}{rgb}{0.5,0.5,0.5}
\definecolor{codepurple}{rgb}{0.58,0,0.82}
\definecolor{backcolour}{rgb}{0.95,0.95,0.92}

\newcommand{\class}[1]{{\colorbox{gray!7}{``#1''}}}

\lstdefinestyle{mystyle}{
    backgroundcolor=\color{backcolour},   
    commentstyle=\color{codegreen},
    keywordstyle=\color{magenta},
    numberstyle=\tiny\color{codegray},
    stringstyle=\color{codepurple},
    basicstyle=\ttfamily\footnotesize,
    breakatwhitespace=false,         
    breaklines=true,                 
    captionpos=b,                    
    keepspaces=true,                 
    numbers=left,                    
    numbersep=5pt,                  
    showspaces=false,                
    showstringspaces=false,
    showtabs=false,                  
    tabsize=2
}

\newif\ifcomments

\commentstrue

\ifcomments
\newcommand{\comments}[1]{#1}
\else
\newcommand{\comments}[1]{}
\fi

\newcommand{\eg}{e.g.\xspace}
\newcommand{\ie}{i.e.\xspace}
\newcommand{\mNPs}{$m$NPs\xspace}
\newcommand{\mNP}{$m$NP\xspace}
\newcommand{\increase}[1]{(\textcolor{ForestGreen}{+#1})}
\newcommand{\increasenoparent}[1]{\textcolor{ForestGreen}{+#1}}
\newcommand{\decrease}[1]{(\textcolor{red}{-#1})}
\newcommand{\decreasenoparent}[1]{\textcolor{red}{-#1}}

\usepackage{pifont}%

\newcommand{\subsec}[1]{\noindent\textbf{#1}~~}
\newcommand{\na}[0]{\textcolor{gray}{n/a}}

\newcommand{\psd}{{PSD}\xspace}
\newcommand{\se}{SS\xspace}

\newcommand{\mybox}[2]{{\color{#1}\fbox{\normalcolor#2}}}
\definecolor{myGreen}{rgb}{0.04,0.58,0.22}
\definecolor{myBlue}{rgb}{0.1,0.4,0.56}

\newcommand\sbullet[1][.5]{\mathbin{\vcenter{\hbox{\scalebox{#1}{$\bullet$}}}}}

\usepackage[capitalize]{cleveref}

\crefname{section}{Sec.}{Secs.}
\Crefname{table}{Table}{Tables}
\Crefname{figure}{Fig.}{Figs.}

\newcommand{\papertitle}{PiC: A Phrase-in-Context Dataset for \\ Phrase Understanding and Semantic Search}
\title{\papertitle}

\author{
    Thang M. Pham $^\dagger$\\
    \texttt{thangpham@auburn.edu} \\
    \And
    Seunghyun Yoon $^\mathsection$ \\
    \texttt{syoon@adobe.com} \\
    \And
    Trung Bui $^\mathsection$\\
    \texttt{bui@adobe.com} \\
    \And
    Anh Nguyen $^\dagger$\\
   \texttt{anh.ng8@gmail.com}
   \AND
    $^\dagger$\textnormal{Auburn University} ~~~~~ $^\mathsection$\textnormal{Adobe Research}
}

\begin{document}

\maketitle

\begin{abstract}
While contextualized word embeddings have been a de-facto standard, learning contextualized \emph{phrase} embeddings is less explored and being hindered by the lack of a human-annotated benchmark that tests machine understanding of phrase semantics given a context sentence or paragraph (instead of phrases alone).
To fill this gap, we propose PiC---a dataset of $\sim$28K of noun phrases accompanied by their contextual Wikipedia pages and a suite of three tasks for training and evaluating phrase embeddings.
Training on PiC improves ranking-models' accuracy and remarkably pushes 
span-selection (\se) models (\ie, predicting the start and end index of the target phrase) near human-accuracy, which is 95\% Exact Match (EM) on semantic search given a query phrase and a passage.
Interestingly, we find evidence that such impressive performance is because the 
\se models learn to better capture the common meaning of a phrase \emph{regardless} of its actual context.
SotA models perform poorly in distinguishing two senses of the same phrase in two contexts ($\sim$60\% EM) and in estimating the similarity between two different phrases in the same context ($\sim$70\% EM).
\end{abstract}

\section{Introduction}

Understanding phrases in context is a key to learning new vocabularies \cite{nagy1985learning,fischer1994learning}, disambiguation \cite{pilehvar-camacho-collados-2019-wic}, and many downstream tasks, including semantic search \cite{finkelstein2001placing}.
Yet, the contextualized \emph{phrase} embeddings \cite{yu-ettinger-2020-assessing} 
in existing systems mostly capture the common meaning of a phrase, \ie without strong dependence on its context \cite{yu-ettinger-2020-assessing}.
While there are \emph{word}-sense disambiguation datasets \cite{edmonds-cotton-2001-senseval,pilehvar-camacho-collados-2019-wic},
no such benchmarks exist for \emph{phrases}.
Existing phrase-similarity benchmarks \cite{pavlick-etal-2015-ppdb,turney2012domain,asaadi-etal-2019-BiRD,paws2019naacl,pawsx2019emnlp} compare phrases alone (without context) and some of them \cite{pavlick-etal-2015-ppdb,paws2019naacl} contain a large, undesired amount ($\sim$15\% to 99\%) of phrase pairs that have lexical overlap (Table~\ref{table:ps_stats}).

Others generated the context for a phrase by querying GPT-2 \cite{wang-etal-2021-phrase} or by retrieving from Wikipedia \cite{yu-ettinger-2020-assessing}. 
Yet, there was no human verification of the realism of generated text \cite{wang-etal-2021-phrase} and no human annotation of how a phrase's meaning changes w.r.t. the context \cite{yu-ettinger-2020-assessing}.
All above drawbacks are limiting the evaluation of phrase understanding.

\begin{figure*}[!hbt]
\centering
\begin{subfigure}{.66\linewidth}
  \centering
  \includegraphics[width=\linewidth]{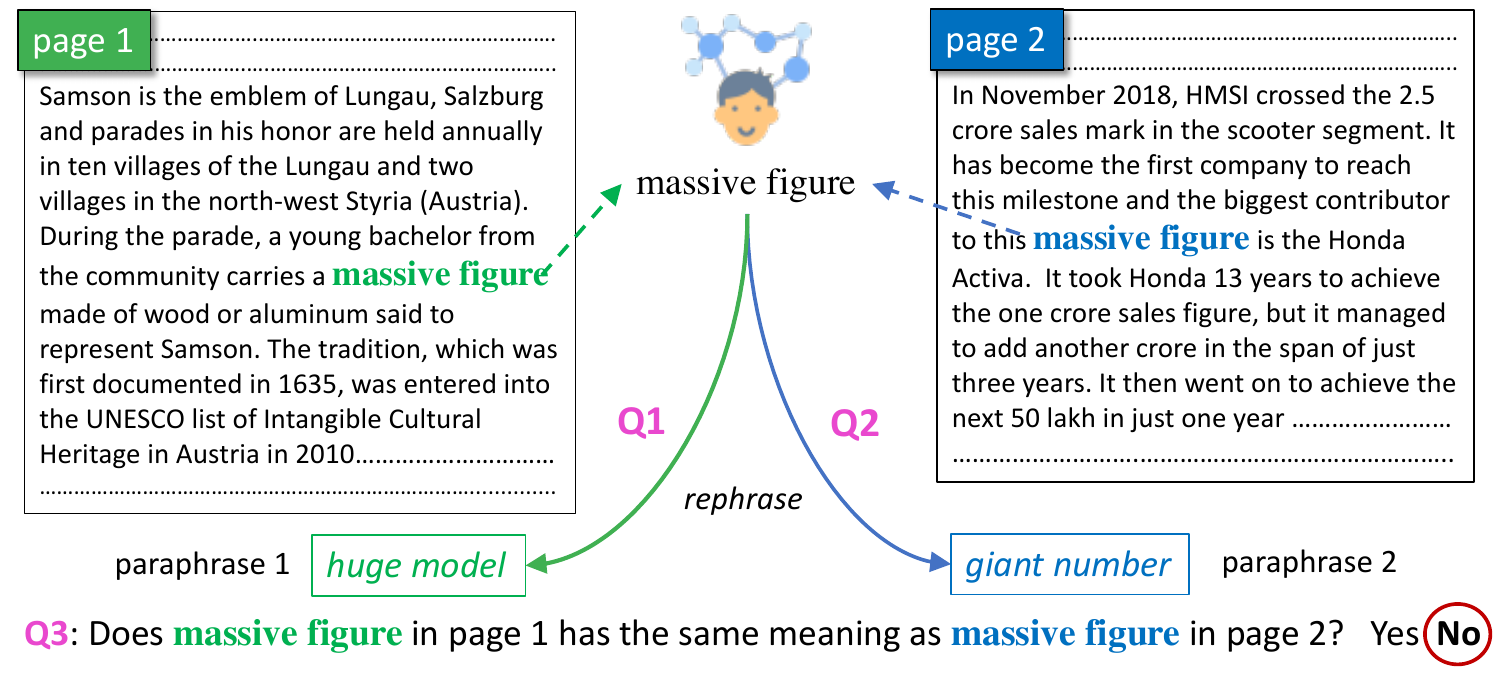}
  \caption{Q1 \& Q2 ask annotators to rephrase ``massive figure'' in page 1 and page 2.
  Q3 asks whether this phrase's meaning is the same in both pages.}
  \label{fig:massive_figure_two_pages}
\end{subfigure}
    \hfill
\begin{subfigure}{.32\linewidth}
    \centering
    \includegraphics[width=\linewidth]{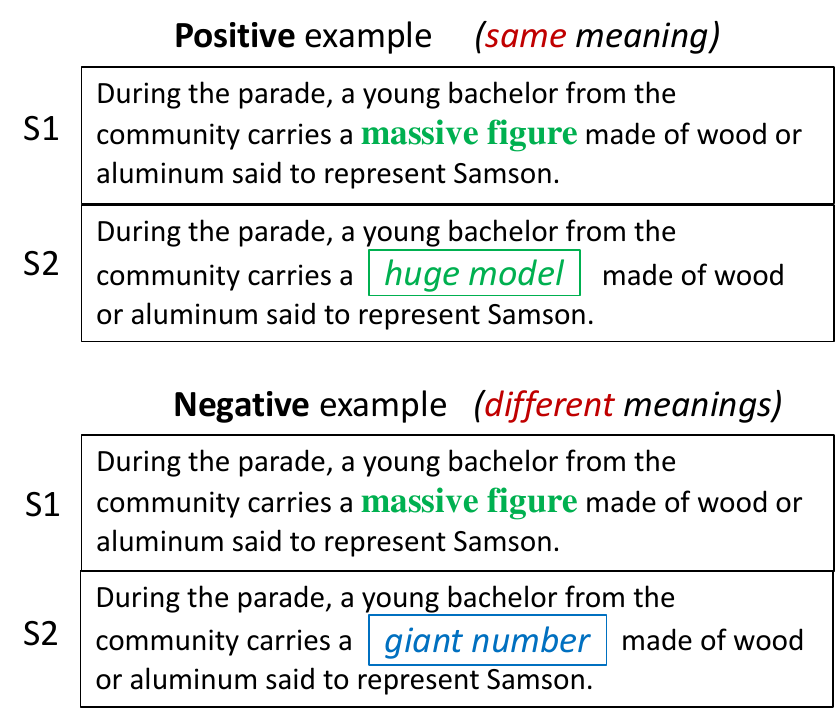}
    \caption{\textbf{PS} positive \& negative examples constructed using page 1 context\\
    (similarly, we repeat for page 2).
    }
    \label{fig:ps_both_examples}
\end{subfigure}

\bigskip

\begin{subfigure}{.36\linewidth}
    \centering
    \includegraphics[width=\linewidth]{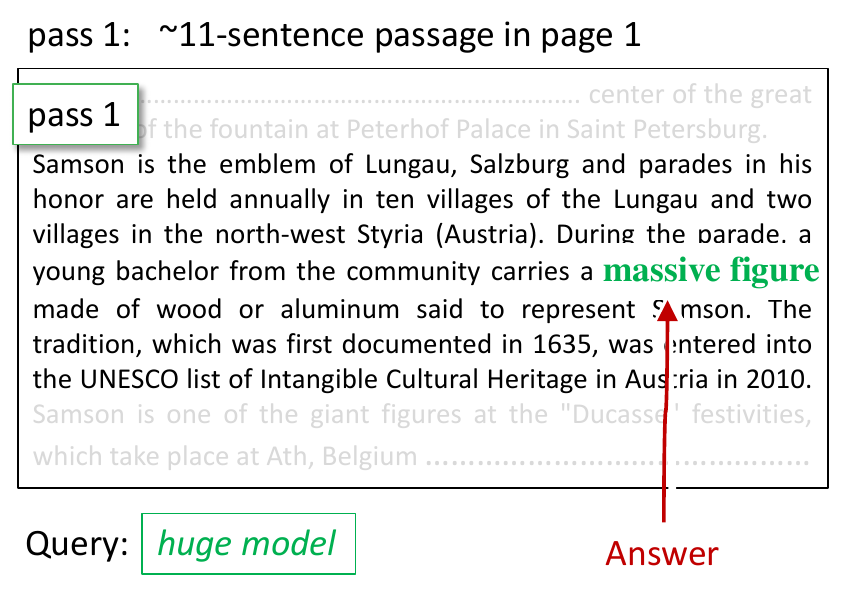}
    \caption{A \textbf{PR-pass} example.}
    \label{fig:pr_pass}
\end{subfigure}
   \hfill
\begin{subfigure}{.24\linewidth}
    \centering
    \includegraphics[width=\linewidth]{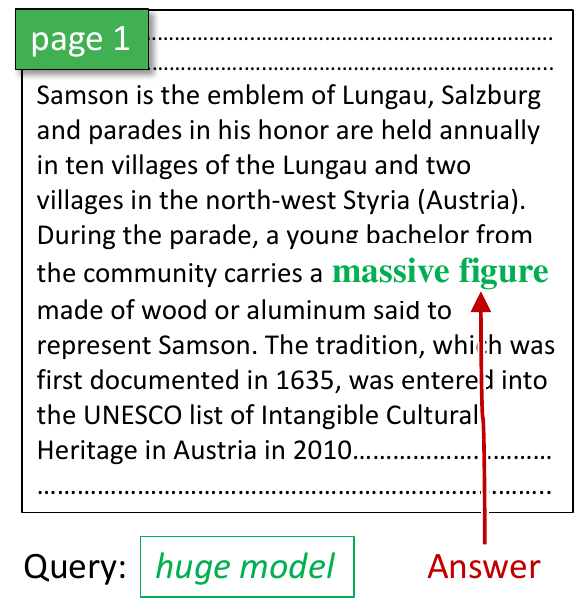}
    \caption{A \textbf{PR-page} example.}
    \label{fig:pr_page}
\end{subfigure}
   \hfill
\begin{subfigure}{.38\linewidth}
    \centering
    \includegraphics[width=\linewidth]{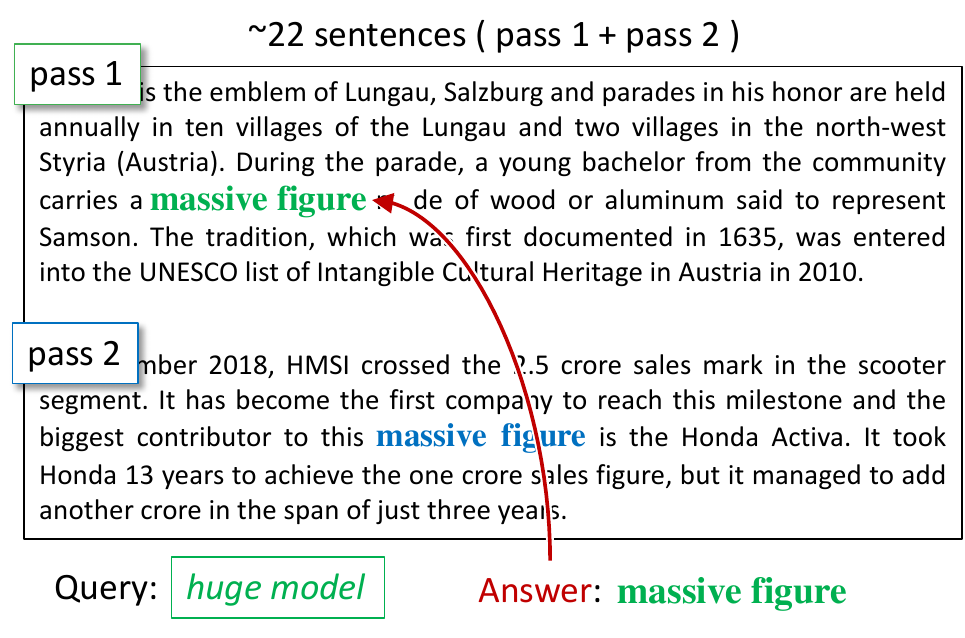}
    \caption{A \textbf{\psd} example.}
    \label{fig:psd}
\end{subfigure}
\caption{Given a phrase, two associated Wikipedia pages, and expert annotations, \ie answers to Q1, Q2, and Q3 (a), we are able to construct \emph{two} pairs of positive and negative examples for PS (b), a PR-pass example (c), a PR-page example (d), and a \psd example \emph{only if} the answer to Q3 is No (e).
}
\label{fig:dataset_construction_summary}
\end{figure*}

To advance the development of contextualized phrase embeddings, we propose Phrase-in-Context (PiC), a suite of three tasks: 
(1) Phrase Similarity (PS), \ie compare the semantic similarity of two phrases in the same context sentence (\cref{fig:ps_both_examples});
(2) Phrase Retrieval (PR), which is divided into PR-pass and PR-page (\cref{fig:pr_pass}--d), \ie from a passage or a Wikipedia page, retrieve a phrase semantically-similar to a given query phrase; 
and (3) Phrase-Sense Disambiguation (\psd),
\ie find the target phrase $\vp$ semantically similar to the query phrase from a 2-paragraph document where $\vp$ appears twice, each time in a different context paragraph that provides a \emph{unique} meaning to $\vp$ (\cref{fig:psd}).
Our $\sim$28K-example dataset is rigorously (a) \emph{annotated} and \emph{verified} by two groups of annotators: linguistics experts on \url{Upwork.com} and non-experts on Amazon Mechanical Turk (MTurk); and then (b) \emph{tested} by models, linguists, and graduate students.
Our contributions are:



\begin{enumerate}
    \item We build PiC\footnote{Dataset, code, and demos are available on \url{https://phrase-in-context.github.io}.}, the first, human-annotated benchmark 
    for evaluating and training contextualized phrase embeddings (\cref{sec:pic}).
    Compared to existing phrase similarity datasets, PS is the first to require models to rely on context.
    
    \item After training on PR-pass, \ie finding a phrase from a passage, \se models perform at a near-human accuracy (92--94\% vs. 95\% EM). 
    They also score high (84--89\% EM) on PR-page, \ie semantic phrase search in a Wikipedia page (Sec.~\ref{sec:results_PR_QA}), suggesting our training set and learned embeddings
    are useful for real-world semantic search.
    
    \item Interestingly, on PR-pass, harnessing these \se models' phrase embeddings in a ranking approach (\ie comparing the similarity between the query and \emph{all} candidate phrases) yields poor accuracy of $\leq$ 59\% EM (Sec.~\ref{sec:results_QA_embeddings}), setting a challenge for future research into learning contextualized phrase embeddings.
    

    \item After training on PR-pass, state-of-the-art (SotA) models perform relatively well on PR-pass and even PR-page but not on \psd (Sec.~\ref{sec:results_PSD}).
    On PS, SotA models perform poorly (below 70\% accuracy) in binary classification of phrase similarity given a context sentence (\cref{sec:result_phrase_similarity}).
    
\end{enumerate}

\section{Related Work}
\label{sec:related_work}

Each of our tasks (PS---phrase similarity; PR---phrase retrieval; and \psd---disambiguation) is related to a separate research area discussed below.

\subsec{Phrase similarity}
First, most existing phrase similarity datasets---\eg PPDB-annotated \cite{wieting-etal-2015-paraphrase}, PPDB-filtered \cite{wang-etal-2021-phrase}, BiRD \cite{asaadi-etal-2019-BiRD}, and PAWS-short \cite{wang-etal-2021-phrase,paws2019naacl}---contain a large percent of instances with lexical overlap between two paired phrases while our PS contains the least percent (5.34\%; \cref{table:ps_stats}).
Second, PS compares each pair of phrases in a context sentence while existing datasets only compare phrases alone (no context).
Third, the phrases in PS are, on average, 2-token long, comparable to that of other datasets (\cref{table:ps_stats}).
Fourth, unlike other datasets, PS contains exclusively \emph{noun-phrases}, the most common phrase type according to Yahoo's search-query statistics \cite{yahooQueryLog} (79.54\%; \cref{appendix_stats_yahoo}) and Adobe (internal Acrobat Pro data not shown).


\begin{table}[ht]
\caption{
Our Phrase Similarity (PS) dataset has a lower 
percent of lexical-overlap instances
and is the \emph{only} human-annotated dataset that provides phrases, each in a context sentence.
}
\label{table:ps_stats}
\centering
\setlength\tabcolsep{1.2pt}
\resizebox{0.5\textwidth}{!}{
\begin{tabular}{lrrrrrrr}
\toprule
& \multicolumn{1}{c}{\textbf{PS}} &
\multicolumn{1}{c}{WiC} &\multicolumn{1}{l}{PPDB-} & \multicolumn{1}{l}{PPDB-} & \multicolumn{1}{r}{BiRD} & \multicolumn{1}{r}{Turney} & \multicolumn{1}{l}{PAWS-} \\
& (\textbf{ours}) &  & annotated & filtered &  &  & short  \\
\cmidrule(){1-8}
\# of All instances & 10,004 & 7,466 & 3,000 & 15,532 & 3,345 & 2,180 & 1,214 \\
\# of Unique phrases & 7,488 & 2,345 & 6,000 & 12,023 & 2,840 & 9,776 & 1,214 \\
Lexical overlap (\%) & 5.34 & 100 & 70.10 & 97.93 & 14.98 & 0 & 99.42 \\
\cmidrule(l{2pt}r{2pt}){1-8}
\multicolumn{2}{l}{Mean length (in tokens)} &  &  \\
    \hspace{1mm} $\sbullet$ phrase$_{1}$ & 2.06 & 1 & 3.67 & 2 & 2 & 2 & 9.52 \\
    \hspace{1mm} $\sbullet$ phrase$_{2}$ & 2.46 & 1 & 3.73 & 2 & 1.49 & 1 & 9.42 \\
    \hspace{1mm} $\sbullet$ context sentence & 22.53 & 8.40 & 0 & 0 & 0 & 0 & 0 \\
\bottomrule
\end{tabular}
}
\end{table}

\subsec{Question answering (QA)}
Our phrase-retrieval tasks---PR and \psd---follow the format of QA datasets except that our queries are \emph{phrases} instead of questions and hence shorter (\cref{table:pr_psd_stats}).
Like SQuAD 1.1 \cite{rajpurkar-etal-2016-squad} and HotpotQA \cite{yang-etal-2018-hotpotqa}, our documents and queries are extracted from Wikipedia articles.
While our PR dataset is $\sim$3.5$\times$ smaller than those two datasets, the paragraph document length in PR-pass and \psd is $\sim$2$\times$ longer than those of SQuAD 1.1 and HotpotQA (\cref{table:pr_psd_stats}).
For our task, intuitively, the longer the document, the harder the task since there would be more candidates a model must compare with the query.


\begin{table}[ht]
\caption{
Our PR-pass, PR-page and \psd datasets are smaller in size compared to common QA datasets and contain shorter queries that are noun phrases instead of questions.
However, our tasks require searching in much longer documents.
}
\label{table:pr_psd_stats}
\centering
\setlength\tabcolsep{1.5pt}
\resizebox{0.48\textwidth}{!}{
\begin{tabular}{lrrrrr}
\toprule
 & \multicolumn{1}{c}{\textbf{PR-pass}} & \multicolumn{1}{c}{\textbf{PR-page}} & 
 \multicolumn{1}{c}{\textbf{\psd}} & 
 \multicolumn{1}{c}{SQuAD 1.1} & \multicolumn{1}{c}{HotpotQA} \\
\cmidrule(){1-6}
All instances & 28,147 & 28,098 & 
4,858
& 98,169 & 105,257 \\
Unique queries/questions & 27,055 & 27,016 & 4,812 & 97,888 & 105,249 \\
Unique answers & 13,458 & 13,423 & 2,314 & 72,469 & 57,259 \\
\cmidrule(l{2pt}r{2pt}){1-6}
Mean length of &  &  &  &  \\
    \hspace{3mm} query (tokens) & 2.42 & 2.42 & 2.45 & 11.42 & 20.03 \\
    \hspace{3mm} answer  (tokens) & 2.17 & 2.17 & 2.07 & 3.46 & 2.35 \\
    \hspace{3mm} sentence (tokens) & 23.22 & 24.08 & 23.00 & 27.62 & 26.77 \\
    \hspace{3mm} document  (sentences) & 10.26 & 119.32 & 20.37 & 5.10 & 4.14 \\
    \hspace{3mm} document  (tokens) & 238.34 & 2,872.73 & 468.48 & 140.92 & 110.72 \\
\bottomrule
\end{tabular}
}
\end{table}


\subsec{Sense disambiguation}
While word-sense disambiguation (WSD) is a long-standing problem in NLP, recently, SotA models have reached super-human accuracy (80\% F$_1$) on the common English WSD \cite{bevilacqua2021recent}.
Interestingly, these high-scoring models still struggle with rare senses that may be outside of the predefined sense inventories or have few training examples \cite{blevins-etal-2021-fews}.
Without the need for predefined senses, WiC \cite{pilehvar-camacho-collados-2019-wic} poses disambiguation as a binary classification task where the goal is to predict whether the same target word in two different sentences carries the same or different meanings.

\subsec{Compared to WiC} PS is also a binary classification task, but with two major differences:
(1) in WiC, the same target word appears in two different sentences while in PS, two different phrases appear in the same context sentence;
(2) PS compares phrases composed of $\geq 2 $ words instead of a single word as in WiC and WSD.
While word senses are defined in WordNet and BabelNet dictionaries \cite{bevilacqua2021recent}, there are no English dictionaries of senses for multi-word noun phrases (\mNPs).
Thus, it is more challenging to acquire and learn the senses of \mNPs, hence the importance of our PiC dataset.
Like WiC, \psd tests disambiguating the meanings of the same $n$-gram in two different contexts.
Yet, \psd is a phrase search task, which involves many more phrase comparisons per example than PS or WiC.

Before the deep learning era, phrase-sense disambiguation was already proposed \cite{carpuat-wu-2007-phrase,carpuat-wu-2007-improving} but only as an auxiliary task for training machine-translation models.
And their phrase senses were not annotated by humans but inferred by performing word-alignment on a bilingual corpus.
Here, our PSD is the first phrase-sense disambiguation task annotated by experts and requires understanding of phrase-senses in a passage.

\section{PiC Dataset Construction}
\label{sec:dataset_construction}


We first collect a set of phrases with context and human annotations.
Then, we derive the examples and labels for three main tasks: PS, PR, and \psd (Fig.~\ref{fig:dataset_construction_summary}).
Our idea is to mine a set of triplets ($\vp$, page$_1$, page$_2$) from Wikipedia where the phrase $\vp$ is a polysemous \mNP that carries two different senses in two Wikipedia pages (\eg, ``massive figure'' means \emph{a large number} in page$_1$ but \emph{a huge physical shape} in page$_2$; \cref{fig:massive_figure_two_pages}).
Then, we ask experts to rephrase $\vp$ into two paraphrases $\vq_1$ and $\vq_2$,  maintaining the two original senses of $\vp$ in page$_1$ and page$_2$, respectively.
The resultant set of 5-tuples ($\vp$, $\vq_1$, $\vq_2$, page$_1$, page$_2$) enables the tests for 
(1) comparing the semantic similarity of two phrases given the same context sentence (PS; \cref{fig:ps_both_examples});
(2) finding a semantically similar phrase in a document (PR-pass \& PR-page; \cref{fig:pr_pass}); 
(3) disambiguating the senses of the same target \mNP given two context paragraphs (\psd; \cref{fig:psd}).

\subsection{Data Collection}
\label{sec:data_collection_method}



As there are no English dictionaries that contain sense inventories for \mNPs, the key \textbf{challenge} to our data collection is to mine \mNPs that have (1) multiple senses; and (2) a Wikipedia context page for each sense.
To do that, we take a Wikipedia dump and perform a \textbf{6-step} procedure that essentially extracts all the \mNPs that occur in more than one Wikipedia page and that contain at least one polysemous word defined in the WiC dataset.
From the triplets of ($\vp$, page$_1$, page$_2$), we programmatically narrow down to $\sim$600K triplets where the context sentence of the \mNP in page$_1$ is the most semantically \emph{dis}similar to the context sentence in page$_2$ (according to SimCSE \citep{gao-etal-2021-simcse}).
We continue filtering down to the top 19,500 triplets where page$_1$ and page$_2$ have \emph{the most} semantically dissimilar lists of Wikipedia categories.
That is, 19,500 triplets are estimated to yield $\sim$15K annotated triplets (the target size based on our budget) after the human annotation process where annotators are allowed to skip the cases they are not confident labeling.
See \cref{sec:appendix_data_collection} for a detailed description of the data collection and \textbf{dataset biases}.


\subsection{Data Annotation}
\label{sec:data_annotation}

Via Upwork, we hire 13 linguistics experts who are native English speakers at a rate of \$30/hour to annotate 15,021 out of 19,500 examples.
For each phrase, we provide Upworkers with a triplet ($\vp$, passage$_1$, passage$_2$) where each passage$_i$ consists of 5 sentences centered at the phrase-containing sentence in the corresponding page$_i$.
We ask them to answer the three below questions (\cref{fig:massive_figure_two_pages}):
\begin{enumerate}[leftmargin=0.5cm,label=Q\arabic*]
    \item Rephrase the target phrase $\vp$ to a paraphrase $\vq_1$ such that its meaning is constant in passage$_1$.
    \item Similarly, rephrase $\vp$ w.r.t. passage$_2$ to obtain a paraphrase $\vq_2$.
    \item Answer Y/N if $\vp$ has the same meaning in both contextual passage$_1$ and passage$_2$.
\end{enumerate}

Upworkers are asked to provide paraphrases that (1) have at least two words and (2) minimize lexical overlap with each other and the target $\vp$.
See the annotation guidelines \cite{annotation_guidelines} and a sample annotation assignment \cite{upwork_samples} given to Upworkers.
After receiving annotations, we use \citet{Language59:online} to automatically find syntactical errors when the paraphrases are replaced by the original target phrase in the original passage and ask Upworkers to fix them.
We also have annotators fix the remaining errors that we find via manual inspection.

\subsection{Annotation Verification}
\label{sec:verify_annotations}

To verify the annotations obtained in \cref{sec:data_annotation} (\ie 2 $\times$ 15,021 = 30,042 paraphrases; and 15,021 Y/N labels), first, we present the same Q1, Q2, and Q3 questions to 1,000 qualified MTurkers and ask whether they agree with the answers by expert annotators in \cref{sec:data_annotation}.
And then, for the cases that the MTurkers disagree with,
we seek second opinions from 5 Upwork experts.
After these two verification rounds, we discard all the examples where Upwork verifiers reject and arrive at the final 28,325 paraphrases and 13,413 Y/N labels (\ie those annotations that \emph{either} an MTurk or Upwork verifier endorses).
See more details in \cref{sec:appendix_verify_annotations}.

The total fee for both MTurk and Upwork combined is around USD 30,000.




\section{Three Phrase Understanding Tasks}
\label{sec:pic}


Using the human-annotated data, we construct three tasks of PS, PR, and \psd (as summarized in \cref{fig:dataset_construction_summary}) for evaluating contextualized phrase-embeddings and semantic-search models.



\subsection{Phrase Similarity (PS)}
\label{sec:phrase_similarity}

PS is a binary classification task that asks whether two \mNPs are semantically similar or not given \emph{the same context} sentence.
The unique challenge of PS is that, \emph{without} context, the two given phrases can be easily interpreted as synonymous.
Yet, in our PS context sentence, the two phrases \emph{may} or may \emph{not} carry distinct meanings (\cref{fig:ps_both_examples}).

\subsec{Construction} From the annotated data, a positive example is a triplet of (an original phrase $\vp$, a paraphrase $\vq_1$, an original page$_1$'s sentence that contains $\vp$).
To create a negative example, from the same triplets, we select only those where the paraphrase $\vq_2$ holds a \emph{different} meaning than $\vq_1$ given the page$_1$ context of $\vq_1$ (\ie, when the answer to Q3 is No; see \cref{fig:ps_both_examples}).
For quality assurance, we also hire three extra Upwork experts to double-check PS annotations (see \cref{sec:appendix_ps_verification}), keeping only examples that at least 2 out of 3 experts endorse.
In total, we obtain 5,002 \emph{negative} examples.
Then, we randomly select 5,002 \emph{positive} examples to form a class-balanced PS dataset.




\subsection{Phrase Retrieval (PR)}
\label{sec:phrase_retrieval}

PR is a task of finding in a given document $\vd$ a phrase $\vp$ that is semantically similar to the given query phrase, which is the paraphrase $\vq_1$ (the answer by annotators to Q1) or $\vq_2$ (the answer to Q2).
We release two versions of PR: \textbf{PR-pass} and \textbf{PR-page}, \ie datasets of triplets (query $\vq_1$, target phrase $\vp$, document $\vd$) where $\vd$ is a random 11-sentence passage that contains $\vp$ (\cref{fig:pr_pass}) or an entire Wikipedia page (\cref{fig:pr_page}).
While PR-pass contains 28,147 examples, PR-page contains slightly fewer examples (28,098) as we remove those examples whose Wikipedia pages coincidentally also contain exactly the query phrase (in addition to the target phrase).
Both datasets are split into $\sim$20K/3K/5K for train/dev/test, respectively.




\subsection{Phrase Sense Disambiguation (\psd)}
\label{sec:phrase_in_context_retrieval}

The task is to find the location of the target phrase $\vp$ where it has a similar meaning to that of the given query $\vq$ in a 2-paragraph document where, by construction, $\vp$ appears exactly twice but only one location is the correct answer (\cref{fig:psd}).

\subsec{Construction} From the verified annotations in \cref{sec:verify_annotations}, there are in total 
4,938
phrases that both annotators and verifiers agree to hold \emph{different} meanings across the two context Wikipedia pages (\ie, ``No'' answer to Q3 in \cref{fig:massive_figure_two_pages}).
To create a \psd example, given a phrase $\vp$ from the above 4,938, we extract two corresponding $\sim$11-sentence paragraphs (from its associated page$_1$ and page$_2$ as in PR-pass) and concatenate them (separated by an empty line) into a single document (\cref{fig:psd}).
Since a \psd example shares a pair of phrases (\emph{query} and \emph{answer}) with one PS \emph{positive} example (\emph{phrase}$_{1}$ and \emph{phrase}$_{2}$), we filter out that \psd example if the corresponding PS example is removed from the additional verification round (\cref{sec:appendix_ps_verification}). 
As the result, we exclude 80 examples and obtain 4,858 examples in total for \psd.




\section{Experiments and Results}
\label{sec:experiments}


We test SotA models on PS, PR-pass, PR-page, and \psd to (1) assess how the models are able to leverage context to improve accuracy; and (2) quantify the headroom for future research.

\subsec{Phrase embeddings}
Besides training and testing SotA BERT-based classifiers, we also test a ranking approach that involves computing the cosine similarity between the query's and each candidate's embedding.
To compute a \emph{contextualized} phrase embedding, following \citealt{yu-ettinger-2020-assessing}, we feed the entire phrase-containing sentence (\eg S$_1$ in \cref{fig:ps_both_examples}) into a model, \eg BERT, and then take the mean pooling of the last-layer embeddings over the words of the given phrase only.
For non-contextualized phrase embeddings, we repeat the same process but input to the model only the phrase (instead of the entire sentence).

\subsec{Models}
\label{sec:models}
We choose SotA models in 
(a) phrase similarity: PhraseBERT \cite{wang-etal-2021-phrase};
(b) sentence similarity: USE-v5 \cite{cer2018universal}, SentenceBERT \cite{reimers-gurevych-2019-sentence}, and SimCSE \cite{gao-etal-2021-simcse});
(c) question-answering: Longformer \cite{beltagy2020longformer}, DensePhrase \cite{lee2021learning}; and 
(d) contextualized embeddings: SpanBERT \cite{joshi2020spanbert} and BERT \cite{devlin2018bert}.

For DensePhrase, we use their Phrase-Encoder (as opposed to the Query-Encoder) to compute phrase embeddings.
USE-v5 is only available via public APIs \cite{USE-v5} that do not support extraction of contextualized phrase embeddings.









\subsection{Phrase Similarity: Contextualized phrase embeddings improve accuracy}
\label{sec:result_phrase_similarity}

\subsec{Q:} \emph{Does incorporating context improve the phrase-similarity accuracy on PS?
}

\subsec{Experiment}
We split the PS dataset 70/10/20 for train/dev/test and test two approaches: (1) using the cosine similarity score between two pre-trained phrase-embeddings (with and without context) to predict phrase similarity; (2) training BERT-based binary classifiers directly using PS training set.
We use 6 backbone BERT models that are all ``base'' versions unless specified otherwise (\cref{table:ps_results}).

\emph{\textbf{Approach 1:} Cosine similarity}~~~
First, we test how \emph{pre-trained} phrase embeddings alone (without finetuning or extra weights) can be leveraged to solve PS.
For each PS example of two phrases, we compute their non-contextualized phrase embeddings and compute their cosine similarity score.
To evaluate the pre-trained embeddings on PS, we follow \citet{arcfacep67:online} and tune the binary-classification threshold $T$ to maximize the training-set accuracy, and then use the same optimal $T$ to report the test-set accuracy.
We repeat the experiment for \emph{contextualized} phrase embeddings.

\emph{\textbf{Approach 2:} BERT-based classifiers}~~~ To complement Approach 1, we test Approach 2, \ie building a binary classifier by adding two extra MLP layers on top of the pre-trained embeddings used in Approach 1.
For a phrase pair, we concatenate the two 768-D phrase embeddings from BERT$_\text{base}$ into a 1,536-D vector, and then place one ReLU layer (256 units) and a 1-output linear classification layer with sigmoid on top.
Following \citet{wang-etal-2021-phrase}, we finetune these models for a maximum of 100 epochs (with early stopping and patience of 10 epochs) on the train set.
See \cref{sec:appendix_experiment_PS} for more training details.

\subsec{Results}
\emph{Without} context, all models perform at $\leq$ 50\% accuracy (\ie the random chance; \cref{table:ps_results}a \& c).
Interestingly, incorporating context information into phrase embeddings substantially improves mean model-accuracy on PS for both Approach 1 (from 50.83\% to 63.43\%; \cref{table:ps_results}b vs. a) and Approach 2 (from 35.40\% to 66.71\%; \cref{table:ps_results}d vs. c), showing evidence that \textbf{PS requires models to rely on context}.
While starting from the same backbone models, Approach 2 yields higher mean accuracy than Approach 1 (\cref{table:ps_results}; 66.71 vs. 63.43), which is expected as Approach 2 models have more capacity and the backbones are allowed to be finetuned on PS.
See Figs.~\ref{fig:ps_example1}--\ref{fig:ps_example4} for qualitative PS predictions from a PhraseBERT-based classifier.

\begin{table}[t]
\caption{Accuracy (\%) of state-of-the-art BERT-based models on the PS test set.
Contextualized phrase embeddings (``Phrase + Ctx'') yield substantially higher performance on PS than non-contextualized embeddings (``Phrase'').
The random baseline is 50\%.
}
\label{table:ps_results}
\centering  
\setlength\tabcolsep{2pt}
\resizebox{0.5\textwidth}{!}{
\begin{tabular}{lll|ll}
\toprule
\multirow{3}{*}{Model} & \multicolumn{2}{c}{\emph{Approach 1:}} & \multicolumn{2}{c}{
 \emph{Approach 2:}} \\
& \multicolumn{2}{c}{\emph{Cosine similarity}} & \multicolumn{2}{c}{
 \emph{BERT-based classifiers}} \\
\cmidrule(l{2pt}r{2pt}){2-3}\cmidrule(l{2pt}r{2pt}){4-5}
& (a) Phrase & (b) Phrase + Ctx & (c) Phrase & (d) Phrase + Ctx \\
\cmidrule(){1-5}
PhraseBERT & 51.75 & 63.40 \increase{11.65} & 33.60 & 66.10 \increase{32.50} \\
\cmidrule(l{2pt}r{2pt}){1-5}
BERT & 51.05 & 64.10 \increase{13.05} & 37.00 & 68.85 \increase{31.85} \\
\cmidrule(l{2pt}r{2pt}){1-5}
SpanBERT & 49.30 & 64.00 \increase{14.70} & 40.15 & 66.85 \increase{26.70} \\
\cmidrule(l{2pt}r{2pt}){1-5}
SpanBERT$_\text{Large}$ & 50.40 & \textbf{66.30} \increase{15.90} & 35.95 & \textbf{69.25} \increase{33.30} \\
\cmidrule(l{2pt}r{2pt}){1-5}
SentenceBERT & 50.35 & 60.30 \increase{9.95} & 31.50 & 62.55 \increase{31.05} \\
\cmidrule(l{2pt}r{2pt}){1-5}
SimCSE & 52.15 & 62.50 \increase{10.35} & 34.20 & 66.65 \increase{32.45} \\
\hline
\cmidrule{1-5}
mean $\pm$ std & 50.83 $\pm$ 1.04 & \textbf{63.43 $\pm$ 1.98} & 35.40 $\pm$ 3.01 & \textbf{66.71 $\pm$ 2.40} \\
\bottomrule
\end{tabular}
}
\end{table}


\subsection{Human Baselines and Upperbound (95\% Exact Match) on Phrase Retrieval}


To interpret the progress of machine phrase-understanding on PR, here, we establish multiple human baselines for both non-experts and linguistics experts (with and without training them).

\subsec{Experiment}
We recruit participants and have them perform one or two tests per person.
A test consists of 20 PR-pass examples.
That is, PR-pass documents are 11-sentence long and are feasible for a person to read in minutes (compared to reading an entire Wikipedia page).
We test three groups: (1) 21 graduate students at our institution (1 test per person); (2) five Upwork experts (1 test per person); and (3) another five Upwork experts (2 tests per person, \ie, for a total of 2 $\times$ 5 = 10 tests).
The students in Group 1 volunteer to help our study unpaid while the Upworkers (Group 2 and 3) are hired using the same procedure as in \cref{sec:data_annotation}.

\begin{table}[t]
\caption{Best \se models reach near the Upperbound (95\%) on PR-pass. 
Yet, ranking models based on phrase embeddings significantly \emph{underperform} \se models.
}
\label{table:human_performance}
\centering
\setlength\tabcolsep{3pt}
\resizebox{0.48\textwidth}{!}{
\begin{tabular}{ll}
\toprule
Accuracy of human groups and models & EM (\%) \\
\cmidrule(){1-2}
Group 1: 20 Non-experts (w/o training) & 73.60 $\pm$ 7.90 \\
Group 2: 05 Experts (w/o training) & 82.00 $\pm$ 12.00 \\
Group 3: 05 Experts (w/ training) & \textbf{90.50} $\pm$ 3.70 \\
 Best human accuracy (4 people)---Upperbound &  \textbf{95.00} $\pm$ 0.00 \\
\cmidrule(l{2pt}r{2pt}){1-2}
 Best untrained, ranking model (BERT) &  47.44  \\
 Best PR-trained, ranking model (PhraseBERT) &  59.02  \\
 Best PR-trained, \se model (Longformer$_\text{Large}$) &  \textbf{94.28} \\
\bottomrule
\end{tabular}
}
\end{table}

\subsec{Results}
First, we find an unsurprising, large gap between non-experts and experts (\cref{table:human_performance}; 73.60\% vs. 82.00\%).
Second, we train experts in Group 3 by having each do a preliminary test and giving them feedback before the real test.
We find the training to substantially boost expert accuracy further (from 82.00\% to 90.50\%).
Importantly, we find the Human Exact Match (EM) Upperbound to be 95\%, \ie the highest scores that 4 people (among all groups) make.
Upon manual inspection of the submissions of these best performers, we find their incorrect answers sometimes partially overlap with the groundtruth or are sometimes reasonable.
In other cases, the best performers find acceptable answers but that do not overlap at all with the groundtruth labels in PR.
That is, we estimate a 5\% of noise in the annotations of PR.




\subsection{Phrase Retrieval: In ranking, context only helps BERT embeddings but not others}
\label{sec:results_PR_ranking}


One way to evaluate the quality of SotA phrase embeddings is by testing:

\subsec{Q:} \emph{How well do phrase embeddings perform in the \textbf{ranking} approach on PR?
}

Ranking is a challenging and meaningful phrase-embedding test because the embedding of the query is compared against that of all phrase candidates (extracted by tokenizing the document), which can include syntactically-incorrect phrases, meaningless phrases or rare phrases.
Such out-of-distribution challenge appears less often in PS or WiC, \ie a binary classification setting.


\subsec{Experiment}
As described in \cref{sec:phrase_retrieval}, the PR train/dev/test splits are 20,147/3K/5K examples and we only use the 5K-example test set to test the models in this ranking experiment (no training).
 We follow \cite{lee-etal-2017-end} for span enumeration to construct a list of candidate phrases, we split each PR document into multiple sentences (using NLTK sentence splitter) and tokenize each sentence into tokens (using NLTK tokenizer) and build an exhaustive list of $n$-grams (here, $n \in \{2,3\}$ only for computational tractability).
For every example, we add the groundtruth phrase (which can be longer than 3 words) to the list of candidates (since we are only interested in testing phrase embeddings, not the phrase extractor).

\subsec{Results}
We report top-$k$ accuracy (for $k$ = 1, 3, 5) and top-5 Mean Reciprocal Rank (MRR@5) on the PR-pass test set in \cref{table:pr_pass_results}a.
First, for most SotA embeddings, incorporating context sentence \emph{hurts} the accuracy (\emph{except for BERT} embeddings).
That is, interestingly, for all BERT embeddings (base and large), the accuracy increases substantially (\increasenoparent{17.64} and \increasenoparent{19.04}; \cref{table:pr_pass_results}) when the one-sentence context is the input.
In contrast, most models that started from BERT but were later finetuned lost the capability to leverage the context information (\eg, PhraseBERT, DensePhrase, and SpanBERT in \cref{table:pr_pass_results}).

Second, the best top-1 accuracy scores on PR-pass for non-contextualized (USE-v5; 43.36\%) and contextualized (BERT; 47.44\%) embeddings are substantially lower than the non-expert baselines (73.60\%; \cref{table:human_performance}) and Human Upperbound (95\%).
Future work is required to learn more robust, phrase embeddings for ranking.
See Figs.~\ref{fig:pr_pass_example}--\ref{fig:pr_page_example} for qualitative examples.

\begin{table*}[t]
\caption{\textbf{Ranking} accuracy (\%) on \textbf{PR-pass} using the state-of-the-art pretrained phrase embeddings. 
See \cref{sec:appendix_quan_results_pr_page} for the results on PR-page.
$\Delta$ (\eg \decreasenoparent{3.62}) denotes the differences between the Top-1 accuracy in the contextualized (``Phrase + Context'') vs. the non-contextualized (``Phrase'') setting.
}
\label{table:pr_pass_results}
\centering
\setlength\tabcolsep{3pt}
\resizebox{0.99\textwidth}{!}{
\begin{NiceTabular}{lcccllccc}
\toprule
 \multirow{2}{*}{Model} & \multicolumn{4}{c}{Phrase} & \multicolumn{4}{c}{Phrase + Context} \\
\cmidrule(l{2pt}r{2pt}){2-5}\cmidrule(l{2pt}r{2pt}){6-9}
& Top-1 & Top-3 & Top-5 & MRR@5~ & Top-1 ($\Delta$) & Top-3 & Top-5 & MRR@5 \\
\cmidrule(){1-9}
\textcolor{brown}{PhraseBERT} \cite{wang-etal-2021-phrase} & 36.62 & 66.96 & 75.90 & 52.20 & 33.00 \decrease{3.62} & 49.60 & 56.70 & 41.90 \\ 
\cmidrule(l{2pt}r{2pt}){1-9}
\textcolor{YellowGreen}{BERT} \cite{devlin2018bert} & 29.80 & 47.90 & 55.40 & 39.50 & \textbf{47.44} \increase{17.64} & \textbf{65.78} & \textbf{73.30} & \textbf{57.30} \\
\cmidrule(l{2pt}r{2pt}){1-9}
\textcolor{Green}{BERT$_\text{Large}$} \cite{devlin2018bert} & 23.76 & 38.52 & 45.40 & 31.70 & \textbf{42.80} \increase{19.04} & \textbf{58.90} & \textbf{64.90} & \textbf{51.30} \\
\cmidrule(l{2pt}r{2pt}){1-9}
\textcolor{BlueViolet}{SpanBERT} \cite{joshi2020spanbert} & 20.88 & 31.04 & 35.20 & 26.40 & 14.40 \decrease{6.48} & 30.46 & 39.80 & 23.40 \\
\cmidrule(l{2pt}r{2pt}){1-9}
\textcolor{Cyan}{SentenceBERT} \cite{reimers-gurevych-2019-sentence} & 22.30 & 50.64 & 60.60 & 36.80 & \textbf{25.14} \increase{2.84} & 39.52 & 46.20 & 32.90 \\
\cmidrule(l{2pt}r{2pt}){1-9}
\textcolor{Purple}{SimCSE} \cite{gao-etal-2021-simcse} & 28.10 & 53.70 & 64.60 & 41.60 & \textbf{32.40} \increase{4.30} & 53.44 & 62.80 & \textbf{43.70} \\
\cmidrule(l{2pt}r{2pt}){1-9}
USE-v5 \cite{cer2018universal} & \textbf{43.36} & \textbf{70.12} & \textbf{78.90} & \textbf{57.30} & \na & \na & \na & \na \\
\cmidrule(l{2pt}r{2pt}){1-9}
{DensePhrase} \cite{lee2021learning} & 32.24 & 51.30 & 60.50 & 42.60 & 31.50 \decrease{0.74} & 46.30 & 53.80 & 39.70 \\
\bottomrule
\end{NiceTabular}
}
\end{table*}


\subsection{Phrase Retrieval: Span-selection models reach near-human accuracy} 
\label{sec:results_PR_QA}


Consistent with \citet{yu-ettinger-2020-assessing}, our ranking results in \cref{sec:results_PR_ranking} reveal that there exists a large headroom for improving both non-contextualized and contextualized phrase embeddings.
Yet, because ranking is a naive approach and \se models \cite{training-qa-hf,devlin2018bert} are the SotA approach on many QA tasks  \cite{rajpurkar-etal-2016-squad}, here we train \se models on the train set of PR-pass and PR-page in order to test:

\subsec{Q:} \emph{How well do SotA semantic-search models perform on PR-pass and PR-page?}

\subsec{Experiment} 
We take the SotA embeddings tested in \cref{sec:results_PR_ranking} and add a linear classification layer on top and finetune each entire classifier on the train set of PR-pass or PR-page for 2 epochs using the default HuggingFace hyperparameters (see \cref{sec:appendix_experiment_PR} for finetuning details).
Following the standard setup of BERT architectures for QA tasks \cite{devlin2018bert}, each \se model predicts the start and end index of the target phrase.
Additionally, since PR-page documents are much longer than a typical QA paragraph (\cref{table:pr_psd_stats}), we also test training Longformer \cite{beltagy2020longformer}, which has a max sequence-length of 4,096, sufficient for an entire Wikipedia page.
We take the models of the smallest dev loss and report their test-set performance in \cref{table:pr_pass_psd_results}.


\subsec{Results} On PR-pass, in contrast to the poor performance of ranking models (\cref{sec:results_PR_ranking}), our PR-pass-trained \se models perform impressively at a near-upperbound level ($\sim$93--94\% EM; \cref{table:pr_pass_psd_results}a) surpassing the accuracy of trained experts (90.50\% EM).
Surprisingly, on PR-page where the documents are substantially longer (around 12$\times$) than the documents of PR-pass, \se models' accuracy only drops slightly (from $\sim$94\% to $\sim$85--89\% EM; \cref{table:pr_pass_psd_results}b).
Note that in a full Wikipedia page of PR-page, there might be phrases that can be considered correct but are \emph{not} labeled groundtruth according to our annotations.
This remarkable result suggests that training on PR-pass can enable high-performing models on real-world semantic search.


\begin{table*}[t]
\caption{\textbf{Test}-set performance (\%) of \textbf{\se models} on PR-pass (a), PR-page (b), and \psd (c).
When trained on PR-pass (a) and PR-page (b), SotA \se models perform well.
However, testing the PR-pass-trained models on \psd shows a significant drop in accuracy (c).
That is, SotA \se models tend to understand a \emph{single sense} of a phrase in context well (high PR-pass, PR-page, and \psd EM scores).
Yet, they are not able to differentiate two senses of the same phrase (\eg, here, PhraseBERT accuracy drops \decreasenoparent{41.27} points between EM+loc vs. EM scores on \psd).
}
\label{table:pr_pass_psd_results}
\centering
\setlength\tabcolsep{4.2pt}
\resizebox{0.99\textwidth}{!}{
\begin{NiceTabular}{lrr|rr|cclc}
\toprule
 \multirow{2}{*}{Model} & \multicolumn{2}{c}{(a) PR-pass} & \multicolumn{2}{c}{(b) PR-page} & \multicolumn{4}{c}{(c) \psd} \\
\cmidrule(l{2pt}r{2pt}){2-3}\cmidrule(l{2pt}r{2pt}){4-5}\cmidrule(l{2pt}r{2pt}){6-9}
& EM & F$_{1}$ & EM & F$_{1}$ & EM & F$_{1}$ & EM+loc & F$_{1}$+loc \\
\cmidrule(){1-9}
\textcolor{brown}{PhraseBERT} \cite{wang-etal-2021-phrase} & 93.42 & 94.97 & 85.24 & 87.19 & 92.98 & 94.08 & 51.67 \decrease{41.31} & 51.83 \\ 
\cmidrule(l{2pt}r{2pt}){1-9}
\textcolor{YellowGreen}{BERT} \cite{devlin2018bert} & 93.26 & 94.65 & 85.64 & 87.77 & 93.50 & 94.57 & 54.84 \decrease{38.66} & 55.07 \\
\cmidrule(l{2pt}r{2pt}){1-9}
\textcolor{Green}{BERT$_\text{Large}$} \cite{devlin2018bert} & 93.64 & 95.16 & 87.36 & 89.52 & 94.67 & 95.57 & 55.43 \decrease{39.24} & 55.61 \\
\cmidrule(l{2pt}r{2pt}){1-9}
\textcolor{BlueViolet}{SpanBERT} \cite{joshi2020spanbert} & 93.50 & 95.02 & 87.28 & 87.66 & 92.26 & 93.30 & 52.20 \decrease{40.06} & 52.34 \\
\cmidrule(l{2pt}r{2pt}){1-9}
\textcolor{Cyan}{SentenceBERT} \cite{reimers-gurevych-2019-sentence} & 93.24 & 94.54 & 84.66 & 86.89 & 93.21 & 94.15 & 52.74 \decrease{40.47} & 52.85 \\
\cmidrule(l{2pt}r{2pt}){1-9}
\textcolor{Purple}{SimCSE} \cite{gao-etal-2021-simcse} & 92.90 & 94.51 & 85.68 & 87.66 & 92.96 & 94.05 & 53.83 \decrease{39.13} & 53.94 \\
\cmidrule(l{2pt}r{2pt}){1-9}
Longformer \cite{beltagy2020longformer} & 94.26 & \textbf{95.58} & \textbf{89.54} & \textbf{91.15} & 96.17 & 96.88 & \textbf{62.72} \decrease{33.45} & \textbf{62.83} \\
\cmidrule(l{2pt}r{2pt}){1-9}
Longformer$_\text{Large}$ \cite{beltagy2020longformer} & \textbf{94.28} & 95.53 & 87.58 & 89.32 & \textbf{96.32} & \textbf{96.91} & 59.72 \decrease{36.60} & 59.82 \\
\hline
\cmidrule{1-9}
mean & 93.56 & 95.00 & 86.92 & 88.85 & 94.01 & 94.94 & \multicolumn{1}{l}{55.39 \decrease{38.62} } & 55.54 \\
$\pm$ std & 0.49 & 0.42 & 1.93 & 1.73 & ~~1.54 & ~~1.36 & \multicolumn{1}{l}{~~3.90} & ~~3.88 \\
\bottomrule
\end{NiceTabular}
}
\end{table*}


\subsection{Phrase Sense Disambiguation: Best models also perform poorly}
\label{sec:results_PSD}


We find that SotA PR-pass-trained \se models reach superhuman accuracy on PR-pass, \ie finding a phrase of the same meaning (\cref{sec:results_PR_QA}).
Yet, PR-pass only tests models' understanding of a \emph{single sense} of the target phrase at a time. 
It is interesting to study: 

\subsec{Q:} \emph{Do PR-pass-trained \se models understand contextualized phrases sufficiently to separate two different senses of the same target phrase?}

\subsec{Experiment}
To do that, here we test the best PR-pass-trained \se models on \psd.
Note that, \psd has the same task format as PR-pass (see \cref{fig:dataset_construction_summary}c--e) except that the document is twice as long and contains \textbf{two occurrences of the same target phrase}. 
We do not test the ranking models as they perform much worse than the \se models in \cref{sec:results_PR_ranking}.

\subsec{Results} 
Although the PR-pass-trained \se models are never trained on \psd, they interestingly frequently find one occurrence of the target phrase (mean of 94.01\% EM; \cref{table:pr_pass_psd_results}c).
However, they mostly locate the \textbf{target phrase in the wrong context passage} with high confidence scores.
That is, if we consider also the correctness of the location of the predicted phrase, their EM+loc\footnote{For a \psd example, if the predicted span does not intersect at all with the groundtruth span, the EM+loc and F$_1$+loc scores would be 0.
If they intersect, the two scores would be equal to EM and F$_1$, respectively.} accuracy drops significantly to an average of 55.39\%.
Also, finetuning on a 2K-example train set of \psd only slightly improves the EM+loc to an average of 64.24\% on a 3K-example \psd test set (\cref{sec:results_PSD-3K}).
Note that we estimate the Human Upperbound on \psd to be 95\%, \ie the same as that of PR-pass.
See qualitative examples and predictions of Longformer (\ie the best model tested) in Figs.~\ref{fig:psd_example_storage_needs}--\ref{fig:psd_example_unrivalled_power}. 

In sum, there is a large headroom for future research on \psd.
\se models are \emph{not} yet capable of leveraging surrounding words to differentiate between two senses of the same phrase.
Interestingly, after training on PR-pass, their contextualized phrase embeddings perform much worse in the ranking experiments on PR-pass (\cref{sec:results_QA_embeddings}).







\section{Discussion and Conclusion}
\label{sec:conclusion}


\subsec{Limitations}
Our dataset is currently limited to multi-word, English noun-phrases.
Furthermore, it is expected to contain around a 5\% error on PR-pass (\ie the best human performance is 95\% EM).
On PR-page, there may be more than one correct target phrase; however, we only label one phrase as the correct answer per document.
We use only phrases that contain at least one WiC word.


While WiC and English WSD rely exclusively on dictionaries \cite{pilehvar-camacho-collados-2019-wic} to obtain word senses and example sentences, our data collection depends on Wikipedia, WiC, \& NLP models and our annotation depends on experts.
In sum, we present PiC, the first 3-task suite for evaluating phrases in context.
\se models can obtain high accuracy on semantic search after training on our PR-pass and PR-page datasets.
Yet, their capability is limited to finding a semantically-similar phrase given a single context that contains the target phrase (in PR-pass).
The results on PS and \psd show that SotA phrase embeddings are still limited in encoding contextualized phrases.
It is interesting future work to improve these models for disambiguating the senses of a phrase in context (PS and \psd).



\subsection*{Acknowledgement}
We thank Qi Li, Peijie Chen, Hai Phan, Giang Nguyen, and Naman Bansal from Auburn University for helpful feedback on the early results. We also thank graduate students from AN's Deep Learning class for helping us with the human study.
We are grateful for the valuable support and feedback from Phat Nguyen. 
AN is supported by NaphCare Foundations, Adobe gifts, and NSF grants (1850117, 2145767).

\clearpage
\bibliography{references}

\begin{thebibliography}{46}
\expandafter\ifx\csname natexlab\endcsname\relax\def\natexlab#1{#1}\fi

\bibitem[{Arici et~al.(2020)Arici, Ceker, and Tutar}]{arici2020multi}
Tarik Arici, Hayreddin Ceker, and Ismail~Baha Tutar. 2020.
\newblock Multi-span question answering using span-image network.
\newblock \emph{preprint}.

\bibitem[{Asaadi et~al.(2019)Asaadi, Mohammad, and
  Kiritchenko}]{asaadi-etal-2019-BiRD}
Shima Asaadi, Saif Mohammad, and Svetlana Kiritchenko. 2019.
\newblock \href {https://doi.org/10.18653/v1/N19-1050} {Big {B}i{RD}: A large,
  fine-grained, bigram relatedness dataset for examining semantic composition}.
\newblock In \emph{Proceedings of the 2019 Conference of the North {A}merican
  Chapter of the Association for Computational Linguistics: Human Language
  Technologies, Volume 1 (Long and Short Papers)}, pages 505--516, Minneapolis,
  Minnesota. Association for Computational Linguistics.

\bibitem[{Beltagy et~al.(2020)Beltagy, Peters, and
  Cohan}]{beltagy2020longformer}
Iz~Beltagy, Matthew~E Peters, and Arman Cohan. 2020.
\newblock \href {https://arxiv.org/abs/2004.05150} {Longformer: The
  long-document transformer}.
\newblock \emph{ArXiv preprint}, abs/2004.05150.

\bibitem[{Bevilacqua et~al.(2021)Bevilacqua, Pasini, Raganato, Navigli
  et~al.}]{bevilacqua2021recent}
Michele Bevilacqua, Tommaso Pasini, Alessandro Raganato, Roberto Navigli,
  et~al. 2021.
\newblock Recent trends in word sense disambiguation: A survey.
\newblock In \emph{Proceedings of the Thirtieth International Joint Conference
  on Artificial Intelligence, IJCAI-21}. International Joint Conference on
  Artificial Intelligence, Inc.

\bibitem[{Bird et~al.(2009)Bird, Klein, and Loper}]{BirdKleinLoper09}
Steven Bird, Ewan Klein, and Edward Loper. 2009.
\newblock \href {https://doi.org/http://my.safaribooksonline.com/9780596516499}
  {\emph{Natural Language Processing with Python: Analyzing Text with the
  Natural Language Toolkit}}.
\newblock O'Reilly, Beijing.

\bibitem[{Blevins et~al.(2021)Blevins, Joshi, and
  Zettlemoyer}]{blevins-etal-2021-fews}
Terra Blevins, Mandar Joshi, and Luke Zettlemoyer. 2021.
\newblock \href {https://doi.org/10.18653/v1/2021.eacl-main.36} {{FEWS}:
  Large-scale, low-shot word sense disambiguation with the dictionary}.
\newblock In \emph{Proceedings of the 16th Conference of the European Chapter
  of the Association for Computational Linguistics: Main Volume}, pages
  455--465, Online. Association for Computational Linguistics.

\bibitem[{Bowman et~al.(2015)Bowman, Angeli, Potts, and
  Manning}]{bowman-etal-2015-large}
Samuel~R. Bowman, Gabor Angeli, Christopher Potts, and Christopher~D. Manning.
  2015.
\newblock \href {https://doi.org/10.18653/v1/D15-1075} {A large annotated
  corpus for learning natural language inference}.
\newblock In \emph{Proceedings of the 2015 Conference on Empirical Methods in
  Natural Language Processing}, pages 632--642, Lisbon, Portugal. Association
  for Computational Linguistics.

\bibitem[{Carpuat and Wu(2007{\natexlab{a}})}]{carpuat-wu-2007-phrase}
Marine Carpuat and Dekai Wu. 2007{\natexlab{a}}.
\newblock \href {https://aclanthology.org/2007.tmi-papers.6} {How phrase sense
  disambiguation outperforms word sense disambiguation for statistical machine
  translation}.
\newblock In \emph{Proceedings of the 11th Conference on Theoretical and
  Methodological Issues in Machine Translation of Natural Languages: Papers},
  Sk{\"o}vde, Sweden.

\bibitem[{Carpuat and Wu(2007{\natexlab{b}})}]{carpuat-wu-2007-improving}
Marine Carpuat and Dekai Wu. 2007{\natexlab{b}}.
\newblock \href {https://aclanthology.org/D07-1007} {Improving statistical
  machine translation using word sense disambiguation}.
\newblock In \emph{Proceedings of the 2007 Joint Conference on Empirical
  Methods in Natural Language Processing and Computational Natural Language
  Learning ({EMNLP}-{C}o{NLL})}, pages 61--72, Prague, Czech Republic.
  Association for Computational Linguistics.

\bibitem[{Cer et~al.(2018)Cer, Yang, Kong, Hua, Limtiaco, John, Constant,
  Guajardo-Cespedes, Yuan, Tar et~al.}]{cer2018universal}
Daniel Cer, Yinfei Yang, Sheng-yi Kong, Nan Hua, Nicole Limtiaco, Rhomni~St
  John, Noah Constant, Mario Guajardo-Cespedes, Steve Yuan, Chris Tar, et~al.
  2018.
\newblock \href {https://arxiv.org/abs/1803.11175} {Universal sentence
  encoder}.
\newblock \emph{ArXiv preprint}, abs/1803.11175.

\bibitem[{Devlin et~al.(2019)Devlin, Chang, Lee, and
  Toutanova}]{devlin2018bert}
Jacob Devlin, Ming-Wei Chang, Kenton Lee, and Kristina Toutanova. 2019.
\newblock \href {https://doi.org/10.18653/v1/N19-1423} {{BERT}: Pre-training of
  deep bidirectional transformers for language understanding}.
\newblock In \emph{Proceedings of the 2019 Conference of the North {A}merican
  Chapter of the Association for Computational Linguistics: Human Language
  Technologies, Volume 1 (Long and Short Papers)}, pages 4171--4186,
  Minneapolis, Minnesota. Association for Computational Linguistics.

\bibitem[{Edmonds and Cotton(2001)}]{edmonds-cotton-2001-senseval}
Philip Edmonds and Scott Cotton. 2001.
\newblock \href {https://aclanthology.org/S01-1001} {{SENSEVAL}-2: Overview}.
\newblock In \emph{Proceedings of {SENSEVAL}-2 Second International Workshop on
  Evaluating Word Sense Disambiguation Systems}, pages 1--5, Toulouse, France.
  Association for Computational Linguistics.

\bibitem[{Finkelstein et~al.(2001)Finkelstein, Gabrilovich, Matias, Rivlin,
  Solan, Wolfman, and Ruppin}]{finkelstein2001placing}
Lev Finkelstein, Evgeniy Gabrilovich, Yossi Matias, Ehud Rivlin, Zach Solan,
  Gadi Wolfman, and Eytan Ruppin. 2001.
\newblock \href {https://doi.org/10.1145/371920.372094} {Placing search in
  context: the concept revisited}.
\newblock In \emph{Proceedings of the Tenth International World Wide Web
  Conference, {WWW} 10, Hong Kong, China, May 1-5, 2001}, pages 406--414.
  {ACM}.

\bibitem[{Fischer(1994)}]{fischer1994learning}
Ute Fischer. 1994.
\newblock Learning words from context and dictionaries: An experimental
  comparison.
\newblock \emph{Applied Psycholinguistics}, 15(4):551--574.

\bibitem[{Gao et~al.(2021)Gao, Yao, and Chen}]{gao-etal-2021-simcse}
Tianyu Gao, Xingcheng Yao, and Danqi Chen. 2021.
\newblock \href {https://doi.org/10.18653/v1/2021.emnlp-main.552} {{S}im{CSE}:
  Simple contrastive learning of sentence embeddings}.
\newblock In \emph{Proceedings of the 2021 Conference on Empirical Methods in
  Natural Language Processing}, pages 6894--6910, Online and Punta Cana,
  Dominican Republic. Association for Computational Linguistics.

\bibitem[{Gebru et~al.(2021)Gebru, Morgenstern, Vecchione, Vaughan, Wallach,
  Iii, and Crawford}]{gebru2021datasheets}
Timnit Gebru, Jamie Morgenstern, Briana Vecchione, Jennifer~Wortman Vaughan,
  Hanna Wallach, Hal~Daum{\'e} Iii, and Kate Crawford. 2021.
\newblock Datasheets for datasets.
\newblock \emph{Communications of the ACM}, 64(12):86--92.

\bibitem[{Group(2022)}]{princeto62:online}
Princeton~NLP Group. 2022.
\newblock princeton-nlp/sup-simcse-roberta-large · hugging face.
\newblock \url{https://huggingface.co/princeton-nlp/sup-simcse-roberta-large}.
\newblock (Accessed on 06/08/2022).

\bibitem[{Honnibal et~al.(2020)Honnibal, Montani, Van~Landeghem, and
  Boyd}]{Honnibal_spaCy_Industrial-strength_Natural_2020}
Matthew Honnibal, Ines Montani, Sofie Van~Landeghem, and Adriane Boyd. 2020.
\newblock \href {https://doi.org/10.5281/zenodo.1212303} {{spaCy:
  Industrial-strength Natural Language Processing in Python}}.
\newblock \emph{https://spacy.io/}.

\bibitem[{Huggingface(2022{\natexlab{a}})}]{superglu53:online}
Huggingface. 2022{\natexlab{a}}.
\newblock super\_glue · datasets at hugging face.
\newblock \url{https://huggingface.co/datasets/super_glue/viewer/wic}.
\newblock (Accessed on 06/08/2022).

\bibitem[{Huggingface(2022{\natexlab{b}})}]{training-qa-hf}
Huggingface. 2022{\natexlab{b}}.
\newblock transformers/examples/pytorch/question-answering at main ·
  huggingface/transformers.
\newblock
  \url{https://github.com/huggingface/transformers/tree/main/examples/pytorch/question-answering}.
\newblock (Accessed on 06/09/2022).

\bibitem[{Joshi et~al.(2020)Joshi, Chen, Liu, Weld, Zettlemoyer, and
  Levy}]{joshi2020spanbert}
Mandar Joshi, Danqi Chen, Yinhan Liu, Daniel~S. Weld, Luke Zettlemoyer, and
  Omer Levy. 2020.
\newblock \href {https://doi.org/10.1162/tacl_a_00300} {{S}pan{BERT}: Improving
  pre-training by representing and predicting spans}.
\newblock \emph{Transactions of the Association for Computational Linguistics},
  8:64--77.

\bibitem[{Kwiatkowski et~al.(2019)Kwiatkowski, Palomaki, Redfield, Collins,
  Parikh, Alberti, Epstein, Polosukhin, Devlin, Lee, Toutanova, Jones, Kelcey,
  Chang, Dai, Uszkoreit, Le, and Petrov}]{kwiatkowski-etal-2019-natural}
Tom Kwiatkowski, Jennimaria Palomaki, Olivia Redfield, Michael Collins, Ankur
  Parikh, Chris Alberti, Danielle Epstein, Illia Polosukhin, Jacob Devlin,
  Kenton Lee, Kristina Toutanova, Llion Jones, Matthew Kelcey, Ming-Wei Chang,
  Andrew~M. Dai, Jakob Uszkoreit, Quoc Le, and Slav Petrov. 2019.
\newblock \href {https://doi.org/10.1162/tacl_a_00276} {Natural questions: A
  benchmark for question answering research}.
\newblock \emph{Transactions of the Association for Computational Linguistics},
  7:452--466.

\bibitem[{LanguageTool(2022)}]{Language59:online}
LanguageTool. 2022.
\newblock Languagetool - online grammar, style \& spell checker.
\newblock \url{https://languagetool.org/}.
\newblock (Accessed on 06/08/2022).

\bibitem[{Lee et~al.(2021)Lee, Sung, Kang, and Chen}]{lee2021learning}
Jinhyuk Lee, Mujeen Sung, Jaewoo Kang, and Danqi Chen. 2021.
\newblock \href {https://doi.org/10.18653/v1/2021.acl-long.518} {Learning dense
  representations of phrases at scale}.
\newblock In \emph{Proceedings of the 59th Annual Meeting of the Association
  for Computational Linguistics and the 11th International Joint Conference on
  Natural Language Processing (Volume 1: Long Papers)}, pages 6634--6647,
  Online. Association for Computational Linguistics.

\bibitem[{Lee et~al.(2017)Lee, He, Lewis, and Zettlemoyer}]{lee-etal-2017-end}
Kenton Lee, Luheng He, Mike Lewis, and Luke Zettlemoyer. 2017.
\newblock \href {https://doi.org/10.18653/v1/D17-1018} {End-to-end neural
  coreference resolution}.
\newblock In \emph{Proceedings of the 2017 Conference on Empirical Methods in
  Natural Language Processing}, pages 188--197, Copenhagen, Denmark.
  Association for Computational Linguistics.

\bibitem[{Nagy et~al.(1985)Nagy, Herman, and Anderson}]{nagy1985learning}
William~E Nagy, Patricia~A Herman, and Richard~C Anderson. 1985.
\newblock Learning words from context.
\newblock \emph{Reading research quarterly}, pages 233--253.

\bibitem[{Pavlick et~al.(2015)Pavlick, Rastogi, Ganitkevitch, Van~Durme, and
  Callison-Burch}]{pavlick-etal-2015-ppdb}
Ellie Pavlick, Pushpendre Rastogi, Juri Ganitkevitch, Benjamin Van~Durme, and
  Chris Callison-Burch. 2015.
\newblock \href {https://doi.org/10.3115/v1/P15-2070} {{PPDB} 2.0: Better
  paraphrase ranking, fine-grained entailment relations, word embeddings, and
  style classification}.
\newblock In \emph{Proceedings of the 53rd Annual Meeting of the Association
  for Computational Linguistics and the 7th International Joint Conference on
  Natural Language Processing (Volume 2: Short Papers)}, pages 425--430,
  Beijing, China. Association for Computational Linguistics.

\bibitem[{PiC(2021{\natexlab{a}})}]{annotation_guidelines}
PiC. 2021{\natexlab{a}}.
\newblock upwork\_annotation\_guidelines.pdf.
\newblock
  \url{https://drive.google.com/file/d/1UsWXvJLzWzuhoRuaCsla3Ljcxp-vM2KK/view?usp=sharing}.
\newblock (Accessed on 06/08/2022).

\bibitem[{PiC(2021{\natexlab{b}})}]{upwork_samples}
PiC. 2021{\natexlab{b}}.
\newblock upwork\_samples.pdf.
\newblock
  \url{https://drive.google.com/file/d/1ume4QeQvEPlwajyeNIQjdH_cCI9w3P6f/view?usp=sharing}.
\newblock (Accessed on 06/08/2022).

\bibitem[{PiC(2022)}]{upwork_verification_samples}
PiC. 2022.
\newblock upwork\_samples.pdf.
\newblock
  \url{https://drive.google.com/file/d/1Lplc_cRla_hbDFzLiJ7ZnO6GOGnRq1gc/view?usp=sharing}.
\newblock (Accessed on 06/08/2022).

\bibitem[{Pilehvar and
  Camacho-Collados(2019)}]{pilehvar-camacho-collados-2019-wic}
Mohammad~Taher Pilehvar and Jose Camacho-Collados. 2019.
\newblock \href {https://doi.org/10.18653/v1/N19-1128} {{W}i{C}: the
  word-in-context dataset for evaluating context-sensitive meaning
  representations}.
\newblock In \emph{Proceedings of the 2019 Conference of the North {A}merican
  Chapter of the Association for Computational Linguistics: Human Language
  Technologies, Volume 1 (Long and Short Papers)}, pages 1267--1273,
  Minneapolis, Minnesota. Association for Computational Linguistics.

\bibitem[{Rajpurkar et~al.(2016)Rajpurkar, Zhang, Lopyrev, and
  Liang}]{rajpurkar-etal-2016-squad}
Pranav Rajpurkar, Jian Zhang, Konstantin Lopyrev, and Percy Liang. 2016.
\newblock \href {https://doi.org/10.18653/v1/D16-1264} {{SQ}u{AD}: 100,000+
  questions for machine comprehension of text}.
\newblock In \emph{Proceedings of the 2016 Conference on Empirical Methods in
  Natural Language Processing}, pages 2383--2392, Austin, Texas. Association
  for Computational Linguistics.

\bibitem[{Reimers and Gurevych(2019)}]{reimers-gurevych-2019-sentence}
Nils Reimers and Iryna Gurevych. 2019.
\newblock \href {https://doi.org/10.18653/v1/D19-1410} {Sentence-{BERT}:
  Sentence embeddings using {S}iamese {BERT}-networks}.
\newblock In \emph{Proceedings of the 2019 Conference on Empirical Methods in
  Natural Language Processing and the 9th International Joint Conference on
  Natural Language Processing (EMNLP-IJCNLP)}, pages 3982--3992, Hong Kong,
  China. Association for Computational Linguistics.

\bibitem[{Team(2021{\natexlab{a}})}]{APICateg56:online}
Wikimedia Team. 2021{\natexlab{a}}.
\newblock Api:categories - mediawiki.
\newblock \url{https://www.mediawiki.org/wiki/API:Categories}.
\newblock (Accessed on 06/08/2022).

\bibitem[{Team(2021{\natexlab{b}})}]{Wikimedi57:online}
Wikimedia Team. 2021{\natexlab{b}}.
\newblock Wikimedia downloads.
\newblock \url{https://dumps.wikimedia.org/}.
\newblock (Downloaded on November 1st, 2021).

\bibitem[{TensorFlow(2022)}]{USE-v5}
TensorFlow. 2022.
\newblock Universal sentence encoder  |  tensorflow hub.
\newblock
  \url{https://www.tensorflow.org/hub/tutorials/semantic_similarity_with_tf_hub_universal_encoder}.
\newblock (Accessed on 08/11/2022).

\bibitem[{Turney(2012)}]{turney2012domain}
Peter~D Turney. 2012.
\newblock Domain and function: A dual-space model of semantic relations and
  compositions.
\newblock \emph{Journal of artificial intelligence research}, 44:533--585.

\bibitem[{Wang et~al.(2021)Wang, Thompson, and Iyyer}]{wang-etal-2021-phrase}
Shufan Wang, Laure Thompson, and Mohit Iyyer. 2021.
\newblock \href {https://doi.org/10.18653/v1/2021.emnlp-main.846}
  {Phrase-{BERT}: Improved phrase embeddings from {BERT} with an application to
  corpus exploration}.
\newblock In \emph{Proceedings of the 2021 Conference on Empirical Methods in
  Natural Language Processing}, pages 10837--10851, Online and Punta Cana,
  Dominican Republic. Association for Computational Linguistics.

\bibitem[{Wieting et~al.(2015)Wieting, Bansal, Gimpel, and
  Livescu}]{wieting-etal-2015-paraphrase}
John Wieting, Mohit Bansal, Kevin Gimpel, and Karen Livescu. 2015.
\newblock \href {https://doi.org/10.1162/tacl_a_00143} {From paraphrase
  database to compositional paraphrase model and back}.
\newblock \emph{Transactions of the Association for Computational Linguistics},
  3:345--358.

\bibitem[{Yahoo(2022{\natexlab{a}})}]{yahooQueryLog}
Yahoo. 2022{\natexlab{a}}.
\newblock Webscope | yahoo labs.
\newblock
  \url{https://webscope.sandbox.yahoo.com/catalog.php?datatype=l&did=66}.
\newblock (Accessed on 08/10/2022).

\bibitem[{Yahoo(2022{\natexlab{b}})}]{Webscope68:online}
Yahoo. 2022{\natexlab{b}}.
\newblock Webscope | yahoo labs.
\newblock
  \url{https://webscope.sandbox.yahoo.com/catalog.php?datatype=l&did=66}.
\newblock (Accessed on 08/10/2022).

\bibitem[{Yang(2022)}]{arcfacep67:online}
Ronghui Yang. 2022.
\newblock arcface-pytorch/test.py at master · ronghuaiyang/arcface-pytorch.
\newblock
  \url{https://github.com/ronghuaiyang/arcface-pytorch/blob/master/test.py}.
\newblock (Accessed on 06/09/2022).

\bibitem[{Yang et~al.(2019)Yang, Zhang, Tar, and Baldridge}]{pawsx2019emnlp}
Yinfei Yang, Yuan Zhang, Chris Tar, and Jason Baldridge. 2019.
\newblock \href {https://doi.org/10.18653/v1/D19-1382} {{PAWS}-{X}: A
  cross-lingual adversarial dataset for paraphrase identification}.
\newblock In \emph{Proceedings of the 2019 Conference on Empirical Methods in
  Natural Language Processing and the 9th International Joint Conference on
  Natural Language Processing (EMNLP-IJCNLP)}, pages 3687--3692, Hong Kong,
  China. Association for Computational Linguistics.

\bibitem[{Yang et~al.(2018)Yang, Qi, Zhang, Bengio, Cohen, Salakhutdinov, and
  Manning}]{yang-etal-2018-hotpotqa}
Zhilin Yang, Peng Qi, Saizheng Zhang, Yoshua Bengio, William Cohen, Ruslan
  Salakhutdinov, and Christopher~D. Manning. 2018.
\newblock \href {https://doi.org/10.18653/v1/D18-1259} {{H}otpot{QA}: A dataset
  for diverse, explainable multi-hop question answering}.
\newblock In \emph{Proceedings of the 2018 Conference on Empirical Methods in
  Natural Language Processing}, pages 2369--2380, Brussels, Belgium.
  Association for Computational Linguistics.

\bibitem[{Yu and Ettinger(2020)}]{yu-ettinger-2020-assessing}
Lang Yu and Allyson Ettinger. 2020.
\newblock \href {https://doi.org/10.18653/v1/2020.emnlp-main.397} {Assessing
  phrasal representation and composition in transformers}.
\newblock In \emph{Proceedings of the 2020 Conference on Empirical Methods in
  Natural Language Processing (EMNLP)}, pages 4896--4907, Online. Association
  for Computational Linguistics.

\bibitem[{Zhang et~al.(2019)Zhang, Baldridge, and He}]{paws2019naacl}
Yuan Zhang, Jason Baldridge, and Luheng He. 2019.
\newblock \href {https://doi.org/10.18653/v1/N19-1131} {{PAWS}: Paraphrase
  adversaries from word scrambling}.
\newblock In \emph{Proceedings of the 2019 Conference of the North {A}merican
  Chapter of the Association for Computational Linguistics: Human Language
  Technologies, Volume 1 (Long and Short Papers)}, pages 1298--1308,
  Minneapolis, Minnesota. Association for Computational Linguistics.

\end{thebibliography}
\bibliographystyle{acl_natbib}



\newcommand{\beginsupplementary}{%
    \setcounter{table}{0}
    \renewcommand{\thetable}{A\arabic{table}}%
    \setcounter{figure}{0}
    \renewcommand{\thefigure}{A\arabic{figure}}%
}

\beginsupplementary%
\appendix


\newcommand{\toptitlebar}{
    \hrule height 4pt
    \vskip 0.25in
    \vskip -\parskip%
}
\newcommand{\bottomtitlebar}{
    \vskip 0.29in
    \vskip -\parskip%
    \hrule height 1pt
    \vskip 0.09in%
}

\newcommand{\suptitle}{Appendix for:\\\papertitle}

Make a 1-column title
\newcommand{\maketitlesupp}{
    \newpage
    \onecolumn
        \null
        \vskip .375in
        \begin{center}
            \toptitlebar
            {\Large \bf \suptitle\par}
            \bottomtitlebar
            \vspace*{24pt}
            {
                \large
                \lineskip=.5em
                \par
            }
            \vskip .5em
            \vspace*{12pt}
        \end{center}
}

\maketitlesupp%




\section{Training models on Phrase Similarity}
\label{sec:appendix_experiment_PS}

\paragraph{Hyperparameters} We train each \emph{BERT-based classifier} for a maximum of 100 epochs with early stopping monitored on validation accuracy (patience of 10 epochs).
We use a batch size of 200 and Adam optimizer with learning rate $\alpha$ = 0.0001, $\beta_1$ = 0.9, $\beta_2$ = 0.999, and $\epsilon = 10^{-8}$.

\paragraph{Training time} On average, with early stopping, training a single model using one V100 GPU takes $\sim$5 and $\sim$8 mins for non-context and context settings, respectively.

\section{Training \se models on Phrase Retrieval}
\label{sec:appendix_experiment_PR}

We finetune each \se model that consists of a linear layer on top of a pretrained model selected in Sec.~\ref{sec:models} to predict the start and end indices of answers (as the common setup in BERT \se models \cite{devlin2018bert,arici2020multi}).
The format of a tokenized input is ``[CLS] query [SEP] document [SEP]'' with maximum sequence length of 4,096 for Longformer$_\text{Base}$ and Longformer$_\text{Large}$ and 512 for the remaining models.
If the document exceeds the maximum sequence length, it is split into smaller features for prediction and thus start and end indices with the highest confidence scores are selected.

\paragraph{Hyperparameters} We follow HuggingFace scheme to finetune the \se models for 2 epochs using Adam optimizer with learning rate $\alpha$ = 0.00003, $\beta_1$ = 0.9, $\beta_2$ = 0.999, $\epsilon = 10^{-8}$.
The batch size varies from 1 to 8 for each model: On one V100 GPU, the ``base'' models can handle 8 examples while the ``large'' BERT models can only fit 2--4 examples into 16GB of memory. 
For Longformer$_\text{Large}$, we use an A100 GPU to feed one PR-page example into the model.
We take the smallest dev-loss models from the training and report their test-set results.

\paragraph{Training time} On average, training a single \se model for 2 epochs using one A100 GPU takes $\sim$20 mins for base models and $\sim$9.5 hours for Longformer$_\text{Large}$.

\section{Data collection}
\label{sec:appendix_data_collection}

From a Wikipedia dump, we perform a 6-step procedure (summarized in \cref{table:summary_data_construction}) for mining a list of \mNPs sorted descendingly by their likelihood of containing multiple senses.
The most polysemous 19,500 \mNPs are then passed to experts for annotation (\cref{sec:data_annotation}) and others for verification (\cref{sec:verify_annotations}).

\paragraph{Step 1: Download Wiki articles} 
We download a Wikipedia dump file \cite{Wikimedi57:online} that contains $\sim$15.78M Wikipedia articles and filter out all empty pages to arrive at $\sim$6.27M non-empty articles. 

\paragraph{Step 2: Extract phrases}



We use NLTK sentence splitter \citep{BirdKleinLoper09} to split each Wikipedia article into multiple sentences.
And then we use SpaCy \cite{Honnibal_spaCy_Industrial-strength_Natural_2020} to extract noun phrases and proper nouns as we do not collect syntactically strict phrases.
For each phrase, we remove all preceding and succeeding stopwords (those among the 179 stopwords in NLTK v3.6.5)
and non-alphanumeric characters.
We remove stopwords because they tend to create more pairs of phrases with lexical overlap, rendering the phrase similarity task easier.
We then remove \emph{unigram} phrases to arrive at $\sim$286.78M \mNPs.
For example, from ``a massive figure'', we changed to ``massive figure'', which would be our final phrase after this step.
For each \mNP, we construct a 3-tuple ({phrase}, {sentence}, {metadata}), \ie the phrase, its container sentence, and metadata for identifying the Wikipedia webpage (hereafter, page).

\paragraph{Step 3: Remove phrases of a single context} 
We further remove all phrases that (1) contain non-ASCII characters (\eg ``phaenná nâsos'', which are non-English); and (2) appear only once, \ie keeping those that occur in multiple sentences since we look for \emph{polysemous} \mNPs, which have multiple senses and contexts.
After this step, $\sim$17.96M phrases remain.

While some phrases with non-ASCII characters are also commonly used in English (\eg, ``d\'ej\`a vu''), we find only 2.48\% of phrases at this stage contain non-ASCII characters, and 29\% of them are common in English.
In short, we are removing only 0.72\% of the English phrases that contain non-ASCII characters in Step 3.


\paragraph{Step 4: Find phrases of polysemous words}
To increase the chance of collecting polysemous \mNPs, we only keep \mNPs that have at least one word in the list of 2,345 unique multiple-sense words of WiC \cite{superglu53:online}, arriving at $\sim$6.5M \mNPs, each appearing in $\geq$ 2 sentences and in $\geq$ 1 Wikipedia pages.
We empirically find that Step 4 is important and substantially increases our chance of finding polysemous \mNPs (compared to skipping Step 4).



\paragraph{Step 5: Find phrases in distinct contexts}

We observe that a \mNP is likely to be polysemous when (a) its context sentences are semantically different; and (b) its context Wikipedia pages are of dissimilar categories (\eg ``massive figure'' in finance vs. history; \cref{fig:psd_example_massive_figure}). 


To implement this filter, we form all possible triplets (phrase, sentence$_1$, sentence$_2$) from the list of context sentences of each \mNP\footnote{For computational tractability, we only keep at most 32 context sentences per \mNP where each sentence's length in words is $\in [5, 25]$.}.
We compute the cosine similarity of two sentences at the CLS embedding space of a SimCSE \citep{gao-etal-2021-simcse} provided on HuggingFace \cite{princeto62:online}.
To find triplets where the two sentences are semantically dissimilar, we keep only the triplets where (sentence$_1$, sentence$_2$) has a low cosine similarity, \ie $\in[-0.3, 0.2]$ and the length difference of the two sentences is < 4 words (as two sentences of substantially different lengths often have a low cosine similarity regardless of their semantic differences).
As the result, there are $\sim$600K triplets remaining after this step.

We further re-rank these $\sim$600K descendingly by the dissimilarity of the lists of Wikipedia categories\footnote{We use the provided Wikipedia API \cite{APICateg56:online} to obtain the categories for each article as the dump file has no category-related information.} of the context pages that contain sentence$_1$ and sentence$_2$.
That is, we treat each Wikipedia page's comma-separated list of categories as an input text to SimCSE and sort the $\sim$600K descendingly by the cosine similarity of the resultant embeddings.

\paragraph{Step 6: Select data for annotation}
Before asking annotators to label our sorted phrases we perform final filtering by removing proper nouns and phrases whose Wikipedia documents contain missing words.

We perform final filtering to ensure the data given to annotators is in a proper format.
That is, from $\sim$600K phrases, we filter down to $\sim$475K phrases by applying two filters: (1) Remove all phrases that are proper nouns (\ie POS tagging returns PROPN) since proper nouns often refer to a single identity and thus unambiguous; (2) Remove all phrases that have a newline character and all phrases whose context Wikipedia page contains missing words (\ie errors in the Wikipedia dump).


As the result, we obtain a list of $\sim$475K phrases sorted by their estimate chance of carrying two different senses.
After manual inspection, we take the top 19,500 triplets of the format (phrase, page$_1$, page$_2$)---\ie a phrase $\vp$ and its two context Wikipedia pages where $\vp$ is the most likely to have two different senses (\eg, see ``massive figure'' in \cref{fig:massive_figure_two_pages})---and hire linguistic experts to annotate them.

\textbf{Our manual inspection} involves taking 1,000 random triplets and manually reading them. 
We find that at least $\sim$30\% of the 1,000-triplet subset contain a polysemous target phrase $\vp$ and two Wikipedia pages that give $\vp$ two unique meanings.
We perform this manual inspection repeatedly throughout the process of inventing and refining the data collection process in order to arrive at the final list of steps as presented in this paper.

\subsection{Biases in the data collection}
\label{sec:biases}

While there are many filtering steps in our data collection above, most of them are data cleaning filters that are typically needed in a regular NLP dataset construction.

We recognize that there are \emph{three key filters} in our system that impose strong biases:

\begin{enumerate}
    \item In Step 4, we use only phrases that contain one word in the  WiC. That is, we find Step 4 to substantially increase our chance of finding triplets with a polysemous target phrase. We have added this note in the Data Collection description. It is possible to remove Step 4, but that would require a larger human annotation effort to reach the same 15K labeled triplets.
    
    \item In Step 5, we rely on SimCSE to find target phrases that are placed in two sentences of dissimilar meanings.
    
    \item In Step 5, we rely on SimCSE to find target phrases that are placed in two Wikipedia pages of distinct topics.

\end{enumerate}








\begin{table}[t]
\caption{
Summary of our 3-stage data construction. 
$\vp$, $s$, $m$ $d$, $q$, $l$ denote target phrase, sentence, metadata, document, query, and label, respectively.
}
\label{table:summary_data_construction}
\centering
\resizebox{0.99\textwidth}{!}{
\begin{tabular}{lrrp{50mm}}
\toprule
 & \multicolumn{1}{c}{Remaining \#} & \multicolumn{1}{c}{Data type} & \multicolumn{1}{c}{Description} \\
\cmidrule(){1-4}
\textbf{\cref{sec:data_collection_method} Data Collection} &  & \\
    \hspace{1mm} Step 1: Download Wiki articles & $\sim$6.27M & articles &  Remove $\sim$9.51M empty articles. \\
    \cmidrule(l{2pt}r{2pt}){2-4}
    \hspace{1mm} Step 2: Extract phrases & $\sim$286.78M & ($\vp$, $s$, $m$) & Extract noun phrases and proper nouns along with their context sentences from Wikipedia articles. \\
    \cmidrule(l{2pt}r{2pt}){2-4}
    \hspace{1mm} Step 3: Remove phrases of a single context & $\sim$17.96M & ($\vp$, $[s_1,...,s_n]$, $m$) & For each phrase, gather all sentences where that phrase is used. \\
    \cmidrule(l{2pt}r{2pt}){2-4}
    \hspace{1mm} Step 4: Find phrases of polysemous words & $\sim$6.5M & ($\vp$, $[s_1,...,s_n]$, $m$) & Filter those phrases that do not contain WiC words. \\
    \cmidrule(l{2pt}r{2pt}){2-4}
    \hspace{1mm} Step 5: Find phrases in distinct contexts &  &  & Sort by $X_i$ and apply filters to find pairs of sentences where their phrase potentially has different meanings. \\
    \hspace{7mm} - Sort and filter by semantic dissimilarity & $\sim$600K & ($\vp$, $s_1$, $s_2$, $m$) & $X_1:$ cosine similarity scores of sentences embeddings. \\
    \hspace{7mm} - Sort by domain dissimilarity & $\sim$600K & ($\vp$, $s_1$, $s_2$, $m$) & $X_2:$ cosine similarity scores of domain embeddings i.e., use categories of each article to get embeddings. \\
    \cmidrule(l{2pt}r{2pt}){2-4}
    \hspace{1mm} Step 6: Select data for expert annotation & 19,500 & ($\vp$, $d_1$, $d_2$) & Remove proper nouns and phrases with missing information and select top 19,500 examples for annotation. \\
\cmidrule(l{2pt}r{2pt}){1-4}
\textbf{\cref{sec:data_annotation} Data Annotations} & 30,042 & ($\vp$, $d$, $q$) & Create a query i.e., paraphrase from the given phrase in each context document. \\
& 15,021 & ($\vp$, $d_1$, $d_2$, $l$) & Create a Yes/No label for each pair of documents. \\
\cmidrule(l{2pt}r{2pt}){1-4}
\textbf{\cref{sec:verify_annotations} Verifying Annotations} & & \\
\hspace{3mm} Round 1: MTurk verifier & 22,496 & ($\vp$, $d$, $q$) & Verify queries and Yes/No label by MTurkers. \\
& 10,043 & ($\vp$, $d_1$, $d_2$, $l$) &  \\
\cmidrule(l{2pt}r{2pt}){2-4}
\hspace{3mm} Round 2: Upwork verifiers & 28,325 & ($\vp$, $d$, $q$) & Verify instances rejected in Round 1. \\
& 13,413 & ($\vp$, $d_1$, $d_2$, $l$) &  \\

\bottomrule
\end{tabular}
}
\end{table}

\section{Statistics for search queries in Yahoo Search Query dataset}
\label{appendix_stats_yahoo}

We analyze 4,496 user queries released in the Yahoo Search Query Log To Entities dataset \citep{Webscope68:online} and use SpaCy tokenizer \cite{Honnibal_spaCy_Industrial-strength_Natural_2020} to classify them into 4 main categories: Noun phrases, verb phrases, URLs and others.
As a result, noun phrases are the most common query type from users with 3,576 queries ($\sim$79.54\%) followed by URLs with 675 queries ($\sim$15.01\%) while verb phrases and other types are less preferred by users.
Moreover, the average length of the real user queries is $\sim$1.60 which is quite close to our PS task with $\sim$2.27.

\begin{table}[H]
\caption{
Statistics of Yahoo queries across different query types.
}
\label{table:yahoo_stats}
\centering  
\setlength\tabcolsep{10pt}
\resizebox{0.45\textwidth}{!}{ 
\begin{NiceTabular}{l|r|r}
\toprule
Query type & \# queries & Percentage (\%) \\
\cmidrule(){1-3}
Noun phrases & 3,576 & 79.54 \\
\cmidrule(l{2pt}r{2pt}){1-3}
Verb phrases & 148 & 3.29 \\
\cmidrule(l{2pt}r{2pt}){1-3}
URLs & 675 & 15.01 \\
\cmidrule(l{2pt}r{2pt}){1-3}
Others & 97 & 2.16 \\
\cmidrule(l{2pt}r{2pt}){1-3}
Total & 4,496 & 100.00 \\
\bottomrule
\end{NiceTabular}
}
\end{table}

\section{Verification of Phrase Similarity}
\label{sec:appendix_ps_verification}



To enhance the quality of the proposed PiC benchmark, we hire three additional Upwork experts to verify the correctness of PS examples where two phrases are supposed to be \emph{non-equivalent} for negative examples (\eg massive figure and giant number in \cref{fig:ps_both_examples}) or \emph{equivalent} for positive examples (\eg massive figure and huge model in \cref{fig:ps_both_examples}), and keep an example if it is endorsed by at least two experts (the rest is discarded from PS). 

\begin{figure*}[h]
  \centering
  \begin{subfigure}[t]{1.0\textwidth}
    \centering
    \includegraphics[width=\linewidth]{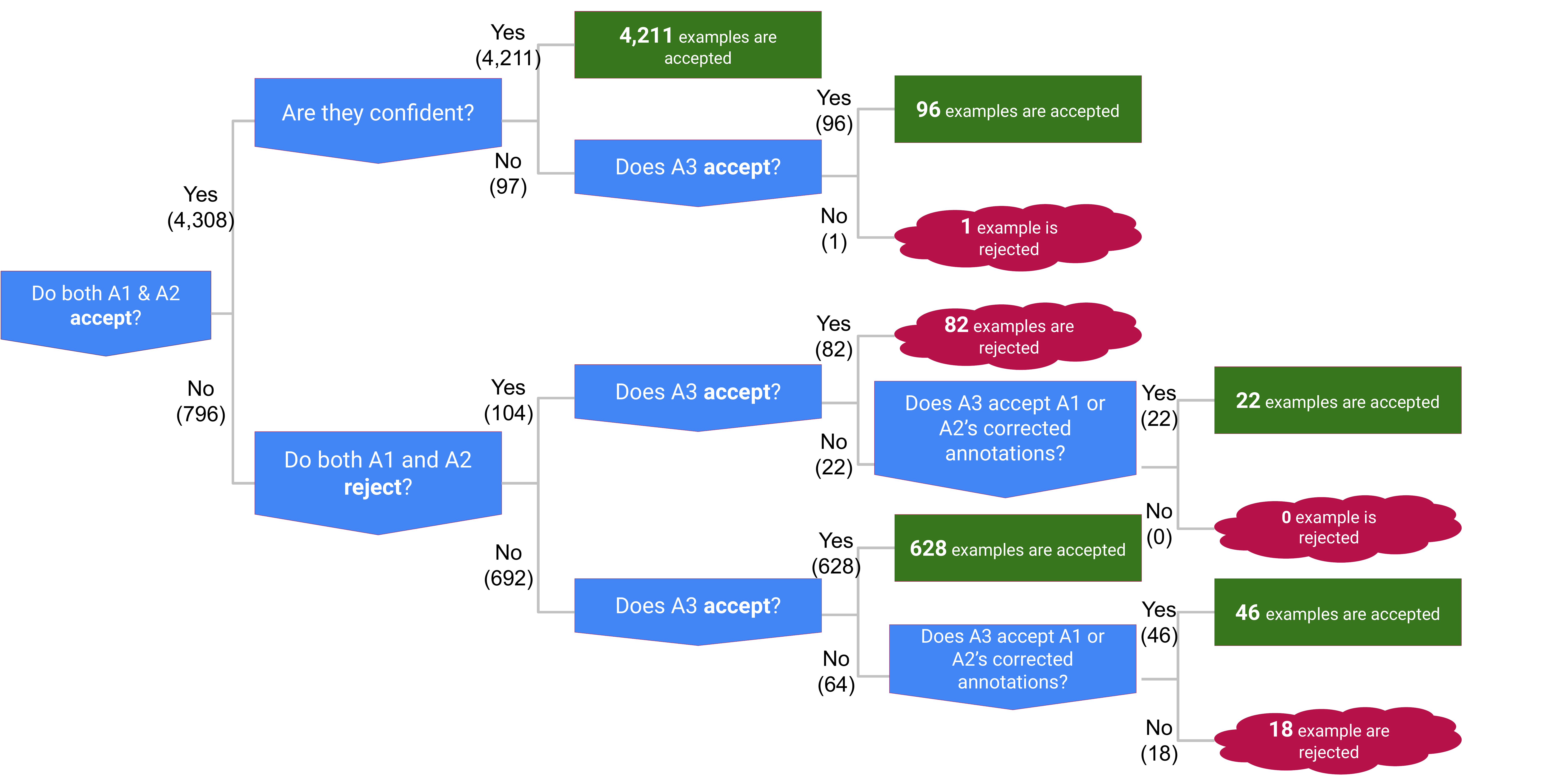}
  \end{subfigure}
  \caption{
  A decision tree describing our verification process for PS that involves three experts.
  Red, green and blue cells represent Reject, Accept decision and Questions.
  The numbers of examples for each branch are shown in parentheses.}
  \label{fig:appendix_decision_tree_ps_negatives}
\end{figure*}

Two Upwork verifiers $A_1 \text{ and } A_2$ start checking 5,104 \emph{negative} examples and the third verifier $A_3$ is responsible for breaking the ties if $A_1 \text{ and } A_2$ disagree with each other (see \cref{fig:appendix_decision_tree_ps_negatives}).
Both $A_1$ and $A_2$ are asked to provide corrections when they do not agree with the labels.
As a result, 4,935 out of 5,104 examples are accepted by pairs of ($A_1$, $A_2$), ($A_1$, $A_3$) or ($A_2$, $A_3$), 68 examples incorrect at first but are modified by either $A_1 \text{ or } A_2$ and endorsed by $A_3$.
In total, we reject 101 negative examples because there are not at least two experts agreeing with the annotations.

\begin{figure}[H]
  \centering
  \begin{subfigure}[t]{1.0\textwidth}
    \centering
    \includegraphics[width=\linewidth]{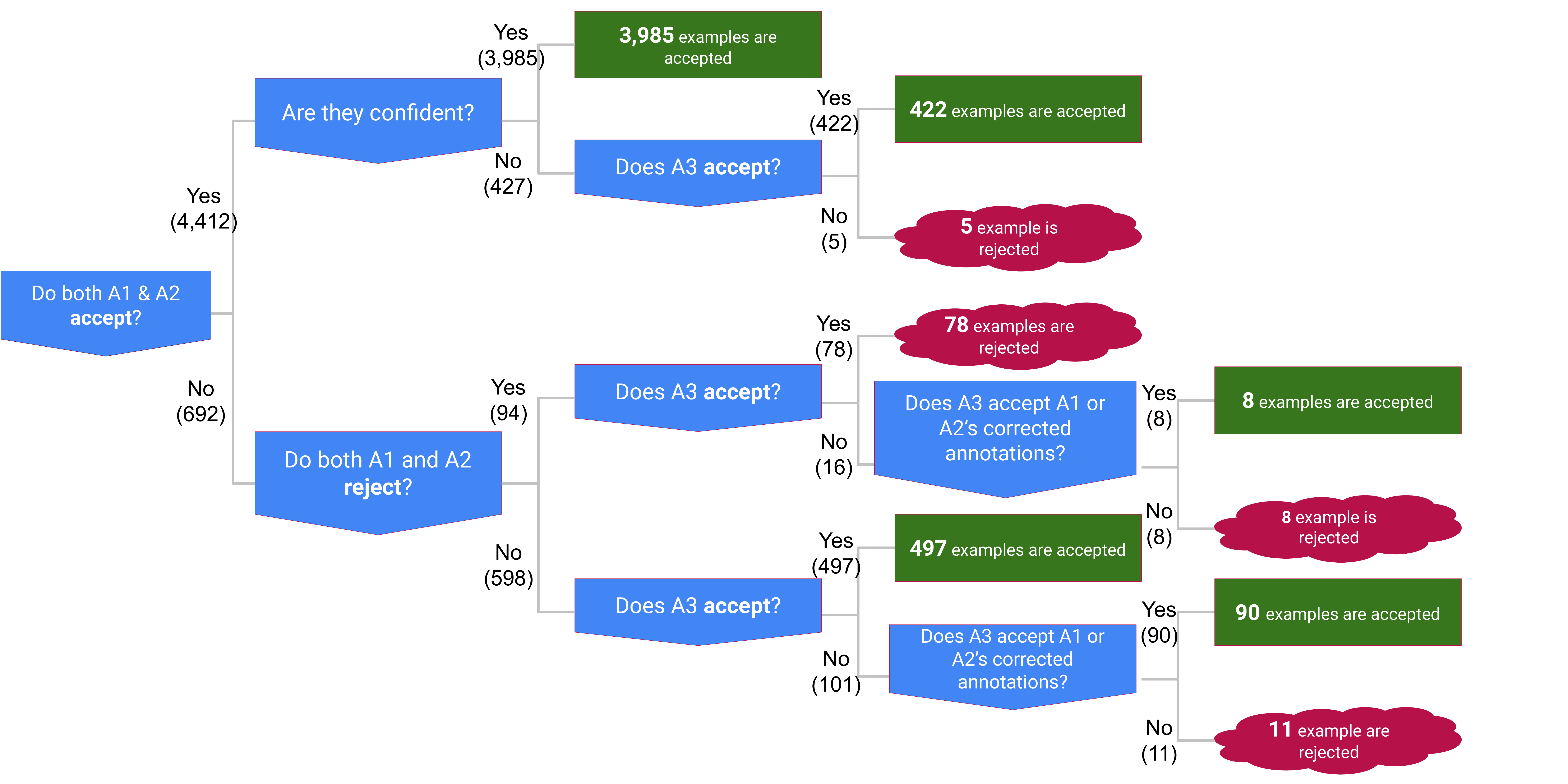}
  \end{subfigure}
  \caption{
  A decision tree describing our verification process for PS that involves three experts.
  Red, green and blue cells represent Reject, Accept decision and Questions.
  The numbers of examples for each branch are shown in parentheses.}
  \label{fig:appendix_decision_tree_ps_positives}
\end{figure}

We repeat the same procedure to verify 5,104 \emph{positive} examples. In sum, we retain 5,002 examples including 4,904 examples accepted by pairs of two verifiers and 98 examples incorrect at first but are modified by either $A_1 \text{ or } A_2$ and endorsed by $A_3$. There are 102 positive examples rejected because there are not at least two experts agreeing with the annotations (\cref{fig:appendix_decision_tree_ps_positives}).

After this verification round, we collect 5,003 \emph{negative} examples and 5,002 \emph{positive} examples. and randomly exclude 1 negative example to make the dataset balance which results in 10,004 examples in total for PS.

\clearpage
\section{Quantitative results on PR-page}
\label{sec:appendix_quan_results_pr_page}

\begin{table*}[h]
\caption{\textbf{Ranking} accuracy (\%) on \textbf{PR-page} using the state-of-the-art pretrained phrase embeddings (a) and those finetuned on PR-pass via \se-style training (b).
}
\label{table:appendix_pr_page_ranking_results}
\centering
\setlength\tabcolsep{5pt}
\resizebox{0.92\textwidth}{!}{
\begin{tabular}{lcccl|lccc}
\toprule
 \multirow{2}{*}{Model} & \multicolumn{4}{c}{Phrase} & \multicolumn{4}{c}{Phrase + Context} \\
\cmidrule(l{2pt}r{2pt}){2-5}\cmidrule(l{2pt}r{2pt}){6-9}
& Top-1 & Top-3 & Top-5 & MRR@5~~ & Top-1 & Top-3 & Top-5 & MRR@5 \\
\cmidrule(){1-9}
\multicolumn{9}{c}{(a) Pre-trained embeddings} \\
\cmidrule(){1-9}
BERT \cite{devlin2018bert} & 20.70 & 34.30 & 41.00 & 28.20 & \textbf{35.40} \increase{14.70} & \textbf{52.10} & \textbf{59.10} & \textbf{44.50} \\
\cmidrule(l{2pt}r{2pt}){1-9}
USE-v5 \cite{cer2018universal} & \textbf{32.20} & \textbf{52.70} & \textbf{60.80} & \textbf{43.20} & \na & \na & \na & \na \\
\hline
\cmidrule(){1-9}
\multicolumn{9}{c}{(b) PR-pass-trained \se models' phrase embeddings} \\
\cmidrule(){1-9}
PhraseBERT \cite{wang-etal-2021-phrase} & \textbf{49.40} & \textbf{69.40} & \textbf{76.70} & \textbf{60.10} & 14.70 & 21.60 & 26.10 & 18.70 \\
\cmidrule(){1-9}
SimCSE \cite{gao-etal-2021-simcse} & 44.20 & 66.60 & 73.50 & 55.70 & 24.60 & 37.80 & 43.20 & 31.70 \\
\bottomrule
\end{tabular}
}
\end{table*}

\section{Finetuning on \psd does not substantially improve accuracy}
\label{sec:results_PSD-3K}

As \psd has only 4,858 examples, we use all examples for testing in \cref{sec:results_PSD} and find the best PR-trained \se models to perform poorly.
To further understand the challenge of \psd, here, we ask:

\subsec{Q:} \emph{How much does training on PR-pass and finetuning on \psd improve accuracy on \psd?}

\paragraph{Experiment}
We take the PR-pass-trained \se models and further finetune them on a subset of \psd to measure how training directly on \psd improves \se models.
We split \psd into 1,438/500/3,000 examples for train/dev/test sets, respectively, and finetune the PR-pass-trained \se models on this \psd train set.
For comparison with the results in \cref{sec:results_PR_QA}, we use the same set of hyperparameters as when finetuning on PR-pass in \cref{sec:results_PR_QA}.
Below, we report the test-set results of the lowest dev-loss models.


\subsec{Results}
On the \psd-3K test set, all models perform poorly at a mean EM score of 55.14\% (\cref{table:psd_3K_results}a; mean).
Interestingly, finetuning the original models using the 1,938 examples (hereafter, \psd-2K) instead of PR-pass decreases accuracy, on average by \decreasenoparent{6.51} points.
An explanation is that 1,438 \psd training examples are too few for the finetuning to be effective.
Indeed, finetuning the PR-pass-trained \se models further on \psd-2K increases the scores for all models by \increasenoparent{9.10} on average (\cref{table:psd_3K_results}c; mean).
The best model is Longformer$_\text{Base}$ \cite{beltagy2020longformer} (\cref{table:psd_3K_results}; 71.10 EM), which is still substantially lower than the human upperbound of 95\%.

\begin{table*}[h]
\caption{Performance of \textbf{\se models} on 3,000 \psd \textbf{test} examples.
(a) and (b) models are \textbf{finetuned} only on PR-pass and 1,938 \psd examples (\psd-2K), respectively. 
(c) models are finetuned on PR-pass first and then finetuned on \psd-2K.
All models are ``base'' unless otherwise specified.
The definitions of EM+loc and F$_1$+loc are in \cref{table:pr_pass_psd_results}'s caption.
}
\label{table:psd_3K_results}
\centering
\setlength\tabcolsep{4.2pt}
\resizebox{0.92\textwidth}{!}{
\begin{NiceTabular}{rrr|lc|lr}
\toprule
 \multirow{1}{*}{Models finetuned on} & \multicolumn{2}{c}{(a) PR-pass} & \multicolumn{2}{c}{(b) \psd-2K} & \multicolumn{2}{c}{(c) PR-pass + \psd-2K} \\
\cmidrule(l{2pt}r{2pt}){2-3}\cmidrule(l{2pt}r{2pt}){4-5}\cmidrule(l{2pt}r{2pt}){6-7}
& EM+loc & F$_{1}$+loc & EM+loc & F$_{1}$+loc & EM+loc & F$_{1}$+loc \\
\cmidrule(){1-7}
PhraseBERT \cite{wang-etal-2021-phrase} & 51.00 & 51.15 & 35.43 \decrease{15.57} & 36.02 & 56.53 ~~\increase{5.53} & 56.81 \\ 
\cmidrule(l{2pt}r{2pt}){1-7}
BERT \cite{devlin2018bert} & 54.53 & 54.75 & 44.33 \decrease{10.20} & 45.28 & 63.83 ~~\increase{9.30} & 64.14 \\ 
\cmidrule(l{2pt}r{2pt}){1-7}
BERT$_\text{Large}$ \cite{devlin2018bert} & 54.77 & 54.99 & 54.07 ~~\decrease{0.70} & 54.82 & 67.13 \increase{12.36} & 67.36 \\
\cmidrule(l{2pt}r{2pt}){1-7}
SpanBERT \cite{joshi2020spanbert} & 52.27 & 52.37 & 44.67 ~~\decrease{7.60} & 45.35 & 69.93 \increase{17.66} & 70.14 \\ 
\cmidrule(l{2pt}r{2pt}){1-7}
SentenceBERT \cite{reimers-gurevych-2019-sentence} & 52.27 & 52.41 & 38.63 \decrease{13.64} & 39.31 & 58.93 ~~\increase{6.66} & 59.21 \\ 
\cmidrule(l{2pt}r{2pt}){1-7}
SimCSE \cite{gao-etal-2021-simcse} & 53.47 & 53.59 & 43.67 ~~\decrease{9.80} & 44.38 & 60.60 ~~\increase{7.13} & 60.80 \\ 
\cmidrule(l{2pt}r{2pt}){1-7}
Longformer \cite{beltagy2020longformer} & \textbf{62.47} & \textbf{62.58} & 61.97 ~~\decrease{0.50} & 62.69 & \textbf{71.10} ~~\increase{8.63} & \textbf{71.30} \\ 
\cmidrule(l{2pt}r{2pt}){1-7}
Longformer$_\text{Large}$ \cite{beltagy2020longformer} & 60.33 & 60.42 & \textbf{66.27} ~\increase{5.94} & \textbf{67.10} & 65.87 ~~\increase{5.54} & 66.10 \\ 
\hline
\cmidrule{1-7}
mean & 55.14 & 55.28 & \multicolumn{1}{l}{48.63 ~~\decrease{6.51} } & 49.37 & \multicolumn{1}{l}{64.24 ~~\increase{9.10}} & 64.48 \\
$\pm$ std & 4.10 & 4.08 & \multicolumn{1}{l}{11.03} & 11.08 & \multicolumn{1}{l}{~~5.23} & 4.13 \\
\bottomrule
\end{NiceTabular}
}
\end{table*}


\clearpage
\section{\se-style training improves \emph{non}-contextualized but not contextualized phrase embeddings}
\label{sec:results_QA_embeddings}

As the \se models trained on PR-pass and PR-page perform impressively (\cref{sec:results_PR_QA}), almost 1.5$\times$ better than the ranking models that are based on pre-trained embeddings, an interesting question is: 

\subsec{Q:} \emph{Does \se training also improve contextualized phrase embeddings?}

This is important to understand because the impressive \se-models' performance gain may come from the extra linear-classification layer (not necessarily from the finetuned embeddings).

\paragraph{Experiment} 
We extract the phrase embeddings (both non-contextualized and contextualized) from the PR-pass-trained \se models from \cref{sec:results_PR_QA} (\ie discarding the classification layer) and test them in the PR-pass ranking experiments (as in \cref{sec:results_PR_ranking}).

\subsec{Results} After finetuning on PR-pass, the \emph{non}-contextualized phrase embeddings improve substantially for most models at an average gain of \increasenoparent{16.61} in top-1 accuracy (\eg, PhraseBERT top-1 accuracy increases from 36.62\% to 59.02\%; \cref{table:pr_pass_results_ranking}b).
This result shows that training on PR-pass improves non-contextualized phrase embeddings.
In stark contrast, the ranking scores of \emph{contextualized} phrase embeddings drop significantly, \decreasenoparent{11.95} points on average (\cref{table:pr_pass_results_ranking}c), compared to before finetuning on PR-pass.

In sum, we are observing a consistent trend that the contextualized phrase embeddings of the original pre-trained BERT (both ``base'' and ``large'') are remarkably beneficial for retrieval (\ie PR).
However, after finetuning, \eg on PR-pass or using other techniques (\eg in PhraseBERT or SentenceBERT), such benefits of leveraging context disappear.
Aligned with \citet{yu-ettinger-2020-assessing}, we find that incorporating context effectively into phrase embeddings is an open research challenge.

\begin{table*}[h]
\caption{\textbf{Ranking} accuracy (\%) on \textbf{PR-pass} using the state-of-the-art pretrained phrase embeddings (a) and those finetuned on PR-pass via \se-style training (b).
See \cref{sec:appendix_quan_results_pr_page} for the results on PR-page.
$\Delta$ (\eg \decreasenoparent{3.62}) denotes the differences between the Top-1 accuracy in the contextualized (``Phrase + Context'') vs. the non-contextualized (``Phrase'') setting.
}
\label{table:pr_pass_results_ranking}
\centering
\setlength\tabcolsep{3pt}
\resizebox{0.88\textwidth}{!}{
\begin{NiceTabular}{lcccllccc}
\toprule
 \multirow{2}{*}{Model} & \multicolumn{4}{c}{Phrase} & \multicolumn{4}{c}{Phrase + Context} \\
\cmidrule(l{2pt}r{2pt}){2-5}\cmidrule(l{2pt}r{2pt}){6-9}
& Top-1 & Top-3 & Top-5 & MRR@5~ & Top-1 ($\Delta$) & Top-3 & Top-5 & MRR@5 \\
\cmidrule(){1-9}
\multicolumn{9}{c}{(a) Pre-trained embeddings} \\
\cmidrule(){1-9}
\textcolor{brown}{PhraseBERT} \cite{wang-etal-2021-phrase} & 36.62 & 66.96 & 75.90 & 52.20 & 33.00 \decrease{3.62} & 49.60 & 56.70 & 41.90 \\ 
\cmidrule(l{2pt}r{2pt}){1-9}
\textcolor{YellowGreen}{BERT} \cite{devlin2018bert} & 29.80 & 47.90 & 55.40 & 39.50 & \textbf{47.44} \increase{17.64} & \textbf{65.78} & \textbf{73.30} & \textbf{57.30} \\
\cmidrule(l{2pt}r{2pt}){1-9}
\textcolor{Green}{BERT$_\text{Large}$} \cite{devlin2018bert} & 23.76 & 38.52 & 45.40 & 31.70 & \textbf{42.80} \increase{19.04} & \textbf{58.90} & \textbf{64.90} & \textbf{51.30} \\
\cmidrule(l{2pt}r{2pt}){1-9}
\textcolor{BlueViolet}{SpanBERT} \cite{joshi2020spanbert} & 20.88 & 31.04 & 35.20 & 26.40 & 14.40 \decrease{6.48} & 30.46 & 39.80 & 23.40 \\
\cmidrule(l{2pt}r{2pt}){1-9}
\textcolor{Cyan}{SentenceBERT} \cite{reimers-gurevych-2019-sentence} & 22.30 & 50.64 & 60.60 & 36.80 & \textbf{25.14} \increase{2.84} & 39.52 & 46.20 & 32.90 \\
\cmidrule(l{2pt}r{2pt}){1-9}
\textcolor{Purple}{SimCSE} \cite{gao-etal-2021-simcse} & 28.10 & 53.70 & 64.60 & 41.60 & \textbf{32.40} \increase{4.30} & 53.44 & 62.80 & \textbf{43.70} \\
\cmidrule(l{2pt}r{2pt}){1-9}
USE-v5 \cite{cer2018universal} & \textbf{43.36} & \textbf{70.12} & \textbf{78.90} & \textbf{57.30} & \na & \na & \na & \na \\
\cmidrule(l{2pt}r{2pt}){1-9}
{DensePhrase} \cite{lee2021learning} & 32.24 & 51.30 & 60.50 & 42.60 & 31.50 \decrease{0.74} & 46.30 & 53.80 & 39.70 \\
\hline
\cmidrule(){1-9}
\multicolumn{9}{c}{(b) PR-pass-trained \se models' phrase embeddings} \\
\cmidrule(){1-9}
\textcolor{brown}{PhraseBERT} \cite{wang-etal-2021-phrase} & \textbf{59.02} & \textbf{81.58} & \textbf{87.90} & \textbf{70.60} & 24.98 \decrease{34.04} & 37.78 & 43.90 & 32.00 \\ 
\cmidrule(l{2pt}r{2pt}){1-9}
\textcolor{YellowGreen}{BERT} \cite{devlin2018bert} & 50.10 & 66.16 & 71.40 & 58.60 & 20.34 \decrease{29.76} & 31.40 & 37.10 & 26.50 \\
\cmidrule(l{2pt}r{2pt}){1-9}
\textcolor{Green}{BERT$_\text{Large}$} \cite{devlin2018bert} & 32.70 & 42.40 & 45.90 & 37.80 & 11.40 \decrease{21.30} & 17.00 & 20.50 & 14.60 \\
\cmidrule(l{2pt}r{2pt}){1-9}
\textcolor{BlueViolet}{SpanBERT}  \cite{joshi2020spanbert} & 15.22 & 22.88 & 26.60 & 19.40 & ~~8.92 \decrease{6.30} & 13.56 & 16.60 & 11.60 \\
\cmidrule(l{2pt}r{2pt}){1-9}
\textcolor{Cyan}{SentenceBERT} \cite{reimers-gurevych-2019-sentence} & 53.14 & 74.86 & 80.70 & 64.20 & 20.12 \decrease{33.02} & 30.04 & 34.90 & 25.60 \\
\cmidrule(l{2pt}r{2pt}){1-9}
\textcolor{Purple}{SimCSE} \cite{gao-etal-2021-simcse} & 50.96 & 76.70 & 83.40 & 64.00 & 37.70 \decrease{13.26} & 52.38 & 58.90 & 45.60 \\
\hline
\cmidrule(){1-9}
\multicolumn{9}{c}{(c) Differences between after vs. before finetuning, \ie the 6 models in (b) vs. those in (a) } \\
\cmidrule(l{2pt}r{2pt}){1-9}
mean differences & \increasenoparent{16.61} &  &  &  &\decreasenoparent{11.95}   &  &  &  \\
\bottomrule
\end{NiceTabular}
}
\end{table*}

\clearpage
\section{Qualitative examples for PS, PR-pass, PR-page and \psd}
\label{sec:appendix_qualitative_examples}

\begin{figure*}[ht]
\centering\small 
\setlength\tabcolsep{2pt} 
\begin{tabular}{|l|p{0.97\linewidth}|}
\hline
\multicolumn{2}{|l|}{\cellcolor{white} \textbf{PS} example. ~ Groundtruth: \class{positive}} \\
\hline
P$_{1}$ & \footnotesize moderate speed \\
\hline
P$_{2}$ & \footnotesize steady pace \\
\hline
S$_{1}$ & \footnotesize Deforestation due to logging and land conversion has likely caused the population to decline at a \colorbox{yellow!50}{\strut moderate speed}. \\
\hline
S$_{2}$ & \footnotesize Deforestation due to logging and land conversion has likely caused the population to decline at a \colorbox{yellow!50}{\strut steady pace}. \\
\hline
\end{tabular}
\caption{
PhraseBERT-based classifier \textcolor{ForestGreen}{\textbf{correctly}} predicts \class{positive} given two phrases P$_{1}$ and P$_{2}$ with and without the presence of context S$_{1}$ and S$_{2}$.
Here, to humans, the phrases are non-polysemous and have the same meaning.
}
\label{fig:ps_example1}
\end{figure*}

\begin{figure*}[ht]
\centering\small 
\setlength\tabcolsep{2pt} 
\begin{tabular}{|l|p{0.97\linewidth}|}
\hline
\multicolumn{2}{|l|}{\cellcolor{white} \textbf{PS} example. ~ Groundtruth: \class{negative}} \\
\hline
P$_{1}$ & \footnotesize greatest emphasis \\
\hline
P$_{2}$ & \footnotesize highest stress \\
\hline
S$_{1}$ & \footnotesize However, the rock art had the \colorbox{yellow!50}{\strut greatest emphasis} on domesticated cattle. \\
\hline
S$_{2}$ & \footnotesize However, the rock art had the \colorbox{yellow!50}{\strut highest stress} on domesticated cattle. \\
\hline
\end{tabular}
\caption{
PhraseBERT-based classifier \textcolor{ForestGreen}{\textbf{correctly}} predicts \class{negative} given two phrases P$_{1}$ and P$_{2}$ with and without the presence of context S$_{1}$ and S$_{2}$.
Here, to humans, the two phrases are non-ambiguously carrying different meanings.
}
\label{fig:ps_example2}
\end{figure*}

\begin{figure*}[ht]
\centering\small 
\setlength\tabcolsep{2pt} 
\begin{tabular}{|l|p{0.97\linewidth}|}
\hline
\multicolumn{2}{|l|}{\cellcolor{white} \textbf{PS} example. ~ Groundtruth: \class{positive}} \\
\hline
P$_{1}$ & \footnotesize unique image \\
\hline
P$_{2}$ & \footnotesize uncommon style \\
\hline
S$_{1}$ & \footnotesize Bayliss has been praised for her \colorbox{yellow!50}{\strut unique image} and tendency to change up songs. \\
\hline
S$_{2}$ & \footnotesize Bayliss has been praised for her \colorbox{yellow!50}{\strut uncommon style} and tendency to change up songs. \\
\hline
\end{tabular}
\caption{
PS case that requires context to determine similarity. 
Without context, a PhraseBERT-based classifier incorrectly thinks P$_{1}$ and P$_{2}$ are different.
Yet, it changes the prediction to \class{positive}, \ie thinking two phrases have the same meaning, when the context is taken into account.
}
\label{fig:ps_example3}
\end{figure*}

\begin{figure*}[ht]
\centering\small 
\setlength\tabcolsep{2pt} 
\begin{tabular}{|l|p{0.97\linewidth}|}
\hline
\multicolumn{2}{|l|}{\cellcolor{white} \textbf{PS} example. ~ Groundtruth: \class{negative}} \\
\hline
P$_{1}$ & \footnotesize permanent post \\
\hline
P$_{2}$ & \footnotesize stable location \\
\hline
S$_{1}$ & \footnotesize His assistant, John Carver took over as caretaker manager, managing one win, but was not considered for the \colorbox{yellow!50}{\strut permanent post}, and left in September 2004. \\
\hline
S$_{2}$ & \footnotesize His assistant, John Carver took over as caretaker manager, managing one win, but was not considered for the \colorbox{yellow!50}{\strut stable location}, and left in September 2004. \\
\hline
\end{tabular}
\caption{
PS case that requires context to determine similarity.
Without context, PhraseBERT-based classifier incorrectly thinks P$_{1}$ and P$_{2}$ carry the same meaning. 
Yet, it correctly changes the prediction to \class{negative} when the context is taken into account.
}
\label{fig:ps_example4}
\end{figure*}

\begin{figure*}[ht]
\centering\small 
\setlength\tabcolsep{2pt} 
\begin{tabular}{|l|p{0.97\linewidth}|}
\hline
\multicolumn{2}{|l|}{\cellcolor{white} \textbf{\psd} example.} \\
\hline
$\vd$ & \footnotesize Bubble memory is a type of non-volatile computer memory that uses a thin film of a magnetic material to hold small magnetized areas, known as "bubbles" or "domains", each storing one bit of data. The material is arranged to form a series of parallel tracks that the bubbles can move along under the action of an external magnetic field. The bubbles are read by moving them to the edge of the material where they can be read by a conventional magnetic pickup, and then rewritten on the far edge to keep the memory cycling through the material. In operation, bubble memories are similar to delay line memory systems. Bubble memory started out as a promising technology in the 1970s, offering memory density of an order similar to hard drives but performance more comparable to core memory while lacking any moving parts. This led many to consider it a contender for a "universal memory" that could be used for all \colorbox{green!50}{\strut storage needs}. The introduction of dramatically faster semiconductor memory chips pushed bubble into the slow end of the scale, and equally dramatic improvements in hard drive capacity made it uncompetitive in price terms. Bubble memory was used for some time in the 1970s and 80s where its non-moving nature was desirable for maintenance or shock-proofing reasons. The introduction of Flash RAM and similar technologies rendered even this niche uncompetitive, and bubble disappeared entirely by the late 1980s. History. Precursors.
\newline\newline
The Inkerman stone, of which the building is made, was mined near Sevastopol and transported by barges. No convenient mooring facilities existed at that time, so the barges had to anchor in the harbor and the load was moved to the shore by boats and then transported to the construction site across the steppe. During the first year of construction, the builders concentrated on the basic structure at the expense of various facilities and decorations. At the end of 1816, the lighthouse looked like a conic 36-metre-high stone tower with a wooden 3.3-metre-high decagonal lantern. The lighthouse became operational in 1817 after its lighting system had been repaired. Three houses were built next to the tower to accommodate the lighthouse personnel and for \colorbox{red!50}{\strut storage needs}. However, cold and humid winters of the Tarkhanut Peninsula, however, made these houses nearly unsuitable for living. In 1862, the lighting system was upgraded, and the spread of light reached 12.4 miles. In 1873, the construction resumed along with cleaning efforts of the surrounding areas. The building was finished and painted white. In 1876, an additional telegraph spot was built near the tower. \\
\hline
$\vq_{1}$ & \footnotesize storehouse purposes ~~ Groundtruth: \colorbox{red!50}{\strut storage needs} \& Prediction: \colorbox{red!50}{\strut storage needs} (confidence: 0.99) \\
\hline
$\vq_{2}$ & \footnotesize data caching ~~~~~~~~~~~~~~ Groundtruth: \colorbox{green!50}{\strut storage needs} \& Prediction: \colorbox{red!50}{\strut storage needs} (confidence: 0.99) \\
\hline
\end{tabular}
\caption{
Given document $\vd$, our Longformer$_\text{Large}$ \se model trained on PR-pass correctly retrieves \colorbox{red!50}{\strut storage needs} in the second paragraph for the query $\vq_{1}$ ``storehouse purposes'' but \emph{fails} to retrieve the answer when the query $\vq_{2}$ is ``data caching''.
The predicted answer for $\vq_{2}$ should be \colorbox{green!50}{\strut storage needs} (\ie in the first passage) since this phrase relates to caching data digitally in computers while \colorbox{red!50}{\strut storage needs} refers to physically storing objects.
}
\label{fig:psd_example_storage_needs}
\end{figure*}

\begin{figure*}[ht]
\centering\small 
\setlength\tabcolsep{2pt} 
\begin{tabular}{|l|p{0.97\linewidth}|}
\hline
\multicolumn{2}{|l|}{\cellcolor{white} \textbf{\psd} example.} \\
\hline
$\vd$ & \footnotesize In the libretto, Delilah is portrayed as a seductive "femme fatale", but the music played during her parts invokes sympathy for her. The 1949 biblical drama "Samson and Delilah", directed by Cecil B. DeMille and starring Victor Mature and Hedy Lamarr in the titular roles, was widely praised by critics for its cinematography, lead performances, costumes, sets, and innovative special effects. It became the highest-grossing film of 1950, and was nominated for five Academy Awards, winning two. According to "Variety", the film portrays Samson as a stereotypical "handsome but dumb hulk of muscle". Samson has been especially honored in Russian artwork because the Russians defeated the Swedes in the Battle of Poltava on the feast day of St. Sampson, whose name is homophonous with Samson's. The lion slain by Samson was interpreted to represent Sweden, as a result of the lion's placement on the Swedish coat of arms. In 1735, C. B. Rastrelli's bronze statue of Samson slaying the lion was placed in the center of the great cascade of the fountain at Peterhof Palace in Saint Petersburg. Samson is the emblem of Lungau, Salzburg and parades in his honor are held annually in ten villages of the Lungau and two villages in the north-west Styria (Austria). During the parade, a young bachelor from the community carries a \colorbox{green!50}{\strut massive figure} made of wood or aluminum said to represent Samson. The tradition, which was first documented in 1635, was entered into the UNESCO list of Intangible Cultural Heritage in Austria in 2010. Samson is one of the giant figures at the "Ducasse" festivities, which take place at Ath, Belgium.
\newline\newline
On September 22, 2015, Honda announced that they had sold over 1 million Activas in five months in the Indian market, from April to August. Honda launched their 5th generation of Honda Activa in 2018, and the sixth-generation Honda Activa 6G have been launched in India with prices starting at 63,912 (ex-showroom, Delhi). Milestones. In April, 2014, "The Economic Times" reported the Honda Activa to be the best selling two wheeler in India, outselling the Hero Splendor. During the month of September 2013, 141,996 Honda Activa scooters were sold, nearly equal to Honda's entire annual sales in North America. The 110cc Activa is the company's biggest seller, by far. It is responsible for over 2,00,000 sales units each month. In November 2018, HMSI crossed the 2.5 crore sales mark in the scooter segment. It has become the first company to reach this milestone and the biggest contributor to this \colorbox{red!50}{\strut massive figure} is the Honda Activa. It took Honda 13 years to achieve the one crore sales figure, but it managed to add another crore in the span of just three years. It then went on to achieve the next 50 lakh in just one year. \\
\hline
$\vq_{1}$ & \footnotesize huge model ~~~~~~~~ Groundtruth: \colorbox{green!50}{\strut massive figure} \& Prediction: \colorbox{red!50}{\strut massive figure} (confidence: 0.99) \\
\hline
$\vq_{2}$ & \footnotesize giant number ~~~~~~ Groundtruth: \colorbox{red!50}{\strut massive figure} \& Prediction: \colorbox{red!50}{\strut massive figure} (confidence: 0.99) \\
\hline
\end{tabular}
\caption{
Given document $\vd$, Longformer$_\text{Large}$ model trained with \se approach on PR-pass correctly retrieves \colorbox{red!50}{\strut massive figure} in the second paragraph for the query $\vq_{2}$ ``giant number'' but \emph{fails} to retrieve the answer when the query $\vq_{1}$ is ``huge model''.
The predicted answer for $\vq_{1}$ should be \colorbox{green!50}{\strut massive figure} in the first passage since this phrase relates to a physical shape instead of a number.
}
\label{fig:psd_example_massive_figure}
\end{figure*}

\begin{figure*}[t]
\centering\small 
\setlength\tabcolsep{2pt} 
\begin{tabular}{|l|p{0.97\linewidth}|}
\hline
\multicolumn{2}{|l|}{\cellcolor{white} \textbf{\psd} example.} \\
\hline
$\vd$ & \footnotesize Eva held ambitions to replace Hortensio Quijano for the 1951 election, although her poor health kept her from this. Nonetheless many were concerned that her agenda would be pushed through. In march of 1951 the government arrested several retired army officers due to their dissent and disapproval of Perón's administration. This raised tensions among the rest of the army, although action did not occur. By September tensions had risen among the military due to the \colorbox{green!50}{\strut unrivalled power} of the Peronist regime. On September 28, 1951, during the election, Menéndez led the military uprising in an attempt to overthrow the government. He led a core of officers, commanding a division, and left Campo de Mayo bound for the Casa Rosada. Resolve for the uprising, especially among the non-commissioned officers and enlisted men, was not strong enough. They were not prepared to fight their own countrymen. The uprising was over as soon as opposition was encountered, almost completely bloodless. Perón admired the loyalty of the troops and pardoned all those involved.
\newline\newline
The design uses a similar standard to the JVX in terms of distortion reduction with crossbraces and 27 cells but that's where the similarity ends. Petra was built from the ground up with entirely new panel shaping and trim. Petra has a highly elliptical planform and very high sweep. NZ Aerosports say she has a high roll rate, a long recovery arc and high maximal glide ratio. She is said to deliver \colorbox{red!50}{\strut unrivalled power} in the turn, plane out and flare. Petra has a long list of World Records, National and International titles to back that up. She had an impressive debut at the PD Big Boy Pants event in July 2011, with Nick Batsch setting a new distance world record of 222.45m (729ft). One month later Nick took out the Pink Open in Klatovy and the FAI World Cup also; first in distance, speed and overall. He also won the 2011 US CP nationals on Petra. Patrick Boulongne came 2nd in the European Championships and 6th overall at the World Cup with Petra in his first competition with her. He went on to win the 2011 French Canopy Piloting Nationals. \\
\hline
$\vq_{1}$ & \footnotesize incomparable energy ~~~~ Groundtruth: \colorbox{red!50}{\strut unrivalled power} \& Prediction: \colorbox{green!50}{\strut unrivalled power} (confidence: 0.99) \\
\hline
$\vq_{2}$ & \footnotesize indomitable strength ~~~~~ Groundtruth: \colorbox{green!50}{\strut unrivalled power} \& Prediction: \colorbox{green!50}{\strut unrivalled power} (confidence: 0.99) \\
\hline
\end{tabular}
\caption{
Given document $\vd$, Longformer$_\text{Large}$ model trained via the \se approach on PR-pass correctly retrieves \colorbox{green!50}{\strut unrivalled power} in the first paragraph for the query $\vq_{2}$ ``indomitable strength'' but \emph{fails} to retrieve the answer when the query $\vq_{1}$ is ``incomparable energy''.
The predicted answer for $\vq_{1}$ should be \colorbox{red!50}{\strut unrivalled power} in the second passage since the second passage changes ``unrivalled power'' meaning to a competition strength instead of military power.
}
\label{fig:psd_example_unrivalled_power}
\end{figure*}

\begin{figure*}[t]
\centering\small 
\setlength\tabcolsep{2pt} 
\begin{tabular}{|l|p{0.97\linewidth}|}
\hline
\multicolumn{2}{|l|}{\cellcolor{white} \textbf{PR-pass} example. \hfill Groundtruth: \colorbox{green!50}{\strut common thought}} \\
\hline
$\vd$ & \footnotesize As the medical corps grew in size there was also specialization evolving. Physicians surfaced that specialized in disease, surgery, wound dressing and even veterinary medicine. Veterinary physicians were there to tend to livestock for agricultural purposes as well as combat purposes. The Cavalry was known for their use of horses in combat and scouting purposes. Because of the type of injuries that would have been commonly seen, surgery was a somewhat common occurrence. Tools such as scissors, knives and arrow extractors have been found in remains. In fact, Roman surgery was quite intuitive, in contrast to \colorbox{green!50}{\strut common thought} of ancient surgery. The Roman military surgeons used a cocktail of plants, which created a sedative similar to modern anesthesia. Written documentation also showed surgeons would use oxidation from a metal such as copper and scrape it into wounds, which provided an antibacterial effect; however, this method was most likely more toxic than providing an actual benefit. Doctors had the knowledge to clean their surgical instruments with hot water after each use. Wounds were dressed, and dead tissue was removed when bandages were changed. \\
\hline
$\vq$ & \footnotesize prevalent theory \\
\hline
R & 0.882~~ \footnotesize \colorbox{green!50}{\strut common thought}\\
& 0.855~~ \footnotesize common thought of\\
& 0.702~~ \footnotesize fact\\
& 0.698~~ \footnotesize to common thought\\ 
& 0.675~~ \footnotesize common occurrence\\
\hline
\end{tabular}
\caption{
A \textbf{ranking} model based on the phrase embeddings of the PR-pass-trained PhraseBERT \se model correctly ranks and retrieves the most semantically relevant answer ``common thought'' as the top-1 prediction in the retrieval list R for the query ``prevalent theory'' in a PR-pass example (which contains a document $\vd$ and a query $\vq$).
}
\label{fig:pr_pass_example}
\end{figure*}

\begin{figure*}[t]
\centering\small 
\setlength\tabcolsep{2pt} 
\begin{tabular}{|l|p{0.97\linewidth}|}
\hline
\multicolumn{2}{|l|}{\cellcolor{white} \textbf{PR-page} example. \hfill Groundtruth: \colorbox{green!50}{\strut continued risk}} \\
\hline
$\vd$ & \footnotesize ... Following a United Nations agreement between Indonesia and Portugal, a UN-supervised referendum held on 30 August 1999 offered a choice between autonomy within Indonesia and full independence. The people of East Timor voted overwhelmingly for independence. An Australian-led and Indonesian-sanctioned peacekeeping force, INTERFET, was sent into the territory to restore order following a violent 'scorched-earth' policy carried out by pro-integration militia and supported by elements of the Indonesian military. In response to Australia's involvement, Indonesia abrogated the 1995 security pact, asserting that Australia's actions in East Timor were inconsistent with 'both the letter and spirit of the agreement'. Official meetings were cancelled or delayed, including the Indonesia-Australia Ministerial Dialogue, which would not reconvene until March 2003. INTERFET was later replaced by a UN force of international police, UNTAET, which formed a detachment to investigate alleged atrocities. "Tampa" affair and the War on Terror. The relationship came under strain in August 2001 during the "Tampa" affair, when Australia refused permission for the Norwegian freighter ship MV "Tampa" to enter Australian waters while carrying Afghan asylum seekers that it had rescued from a distressed fishing vessel in international waters. The Indonesian Search and Rescue Agency did not immediately respond to requests from Australia to receive the vessel. When the ship entered Australian territorial waters after being refused permission, Australia attempted without success to persuade Indonesia to accept the asylum seekers. Norway also refused to accept the asylum seekers and reported Australia to international maritime authorities. The incident prompted closer coordination between Indonesian and Australian authorities, including regional conferences on people smuggling, trafficking in persons and other transnational crime. In 2002, a terrorist attack in Kuta, Bali killed 202 people, including 88 Australians, and injured a further 240. Jemaah Islamiyah, a violent Islamist group, claimed responsibility for the attack, allegedly in retaliation for Australia's support for East Timorese independence and the War on Terror. A subsequent attack in 2005 resulted in the deaths of a further 20 people, including 15 Indonesians and 4 Australians. The 2003 Marriott Hotel bombing was also perceived as targeted at Western interests in Indonesia; Al Qaeda claimed the attack was carried out by a Jemaah Islamiyah suicide bomber in response to actions of the United States and its allies, including Australia. A 2004 attack on the Australian embassy in Jakarta by Jemaah Islamiyah resulted in the deaths of nine Indonesians. The following year, Indonesian diplomatic and consular premises in Australia received a number of hoax and threat messages. Since then, both the United States and Australian governments have issued warnings against travel to Indonesia, advising their citizens of a \colorbox{green!50}{\strut continued risk} of attacks. These incidents prompted greater cooperation between law enforcement agencies in the two countries, building on a 1999 agreement on drug trafficking and money laundering. The Australian Federal Police's Jakarta Regional Cooperation Team provided assistance to the Indonesian National Police, and has contributed to the Jakarta Centre for Law Enforcement Cooperation. This relationship has attracted criticism, particularly following the arrest and sentencing of the Bali Nine, a group of nine Australians arrested in Denpasar while attempting to smuggle heroin from Indonesia to Australia. The 2005 conviction of Schapelle Corby for attempting to smuggle drugs to Bali also attracted significant attention in the Australian media. The 2004 Indian Ocean earthquake prompted a significant humanitarian response from Australia, including a \$1 billion aid package from the federal government, a further \$17.45 million contribution from state and territory governments, and the commitment of 900 Australian Defence Force personnel to relief efforts in northern Sumatra and Aceh. A telethon broadcast on Australia's three major commercial television networks called "" generated pledges of more than \$10 million, contributing to total private aid of \$140 million. The Eighth "Australia-Indonesia Ministerial Forum" (AIMF) was held in Bali on 29 June 2006 and was attended by five Australian and eleven Indonesian ministers. A key outcome was support for the conclusion of a security agreement, later realised as the Lombok Agreement, providing a framework for the development of the security relationship by the end of 2006 on defence, law enforcement, counter-terrorism, intelligence, maritime security, aviation safety, WMD non-proliferation, and bilateral nuclear cooperation for peaceful purposes. Australia-Indonesia-East Timor Trilateral Ministerial Meetings occurred three times to September 2006. Recent relations. 2010 President Susilo Bambang Yudhoyono visited Australia in April 2010, and became the second Indonesian leader to address federal parliament: Finally, I look forward to a day in the near future. The day when policy makers, academicians, journalists and other opinion leaders all over the world take a good look at the things we are doing so well together. And they will say: these two used to be worlds apart. But they now have a fair dinkum of a partnership. ... \\
\hline
$\vq$ & \footnotesize sustained threat \\
\hline
R & 0.830~~\footnotesize threat . \\
& 0.802~~\footnotesize potential threat \\
& 0.800~~\footnotesize threat reached \\
& 0.787~~\footnotesize threat as \\ 
& 0.787~~\footnotesize threat to \\
\hline
\end{tabular}
\caption{
A \textbf{ranking} model based on the non-contextualized embeddings of USE-v5 fails to retrieve the correct answer ``continued risk'' for the query ``sustained threat'' in the PR-page example (which contains a document $\vd$ and a query $\vq$).
The top-5 phrases retrieved (R) contains the word ``threat'' but have no identifier conveying the ``continued'' or `sustained'' sense.
Here, the Wikipedia page is truncated to fit into a single manuscript page.
}
\label{fig:pr_page_example}
\end{figure*}

\clearpage
\section{Verifying annotations}
\label{sec:appendix_verify_annotations}


There are two common methods for evaluation of dataset quality: 
(1) Verify only a small, random subset \citep{pilehvar-camacho-collados-2019-wic} to estimate the quality of the full dataset or (2) verifying the entire dataset with multiple annotators and use the inter-annotator agreement (IAA) to control quality \citep{bowman-etal-2015-large, kwiatkowski-etal-2019-natural}.
The first approach for approximation is budget-friendly but it remains unknown whether the rest of examples are at high quality, while IAA is more desired but annotating thousands of instances can be prohibitively slow and costly.

We propose a \textbf{hybrid approach} to evaluate (leveraging both linguistic experts and non-experts) and ensure high quality for 30,042 queries and 15,021 Yes/No answers at lower cost compared to IAA via two rounds:
\begin{enumerate}
    \item First, we ask around 1,000 highly qualified freelancers on Amazon Mechanical Turk (MTurk verifiers) to verify whether the \emph{query} annotated by our Upwork annotators is interchangeable \ie has the same meaning with the given \emph{phrase} in \emph{paragraph}.
To verify Yes/No answers, MTurk verifiers need to read two short paragraphs containing the same phrase like Upwork annotators to make decisions.
We do not show answers to the MTurk verifiers to avoid biases.

\item Second, we continue hiring 5 Upwork verifiers who are writing experts to double-check those instances rejected by MTurk verifiers from the previous round and only discard an example if the Upwork verifiers agree with MTurk verifiers.
\end{enumerate}


\subsection{Round 1: Verification by MTurk non-experts}
\label{sec:crowdsourcing_evaluation}

We use AMT platform to recruit more than 1,000 MTurk verifiers.
Also, we use Gorrila (\url{gorilla.sc}) to develop user interface to collect answers from participants because (1) Gorilla provides easy-to-use tools to build graphical interface, (2) it is straightforward to monitor and discard results from unqualified participants and (3) we can easily share the experiment with MTurk verifiers via a link.
Per 30 verified answers in around $\sim$20 minutes, the verification process costs us \$5.6 (AMT fees included) and 1 token to Gorilla to a single MTurk verifier.

\begin{figure*}
\begin{subfigure}{.5\textwidth}
  \centering
  \mybox{YellowGreen}{\includegraphics[width=.95\linewidth]{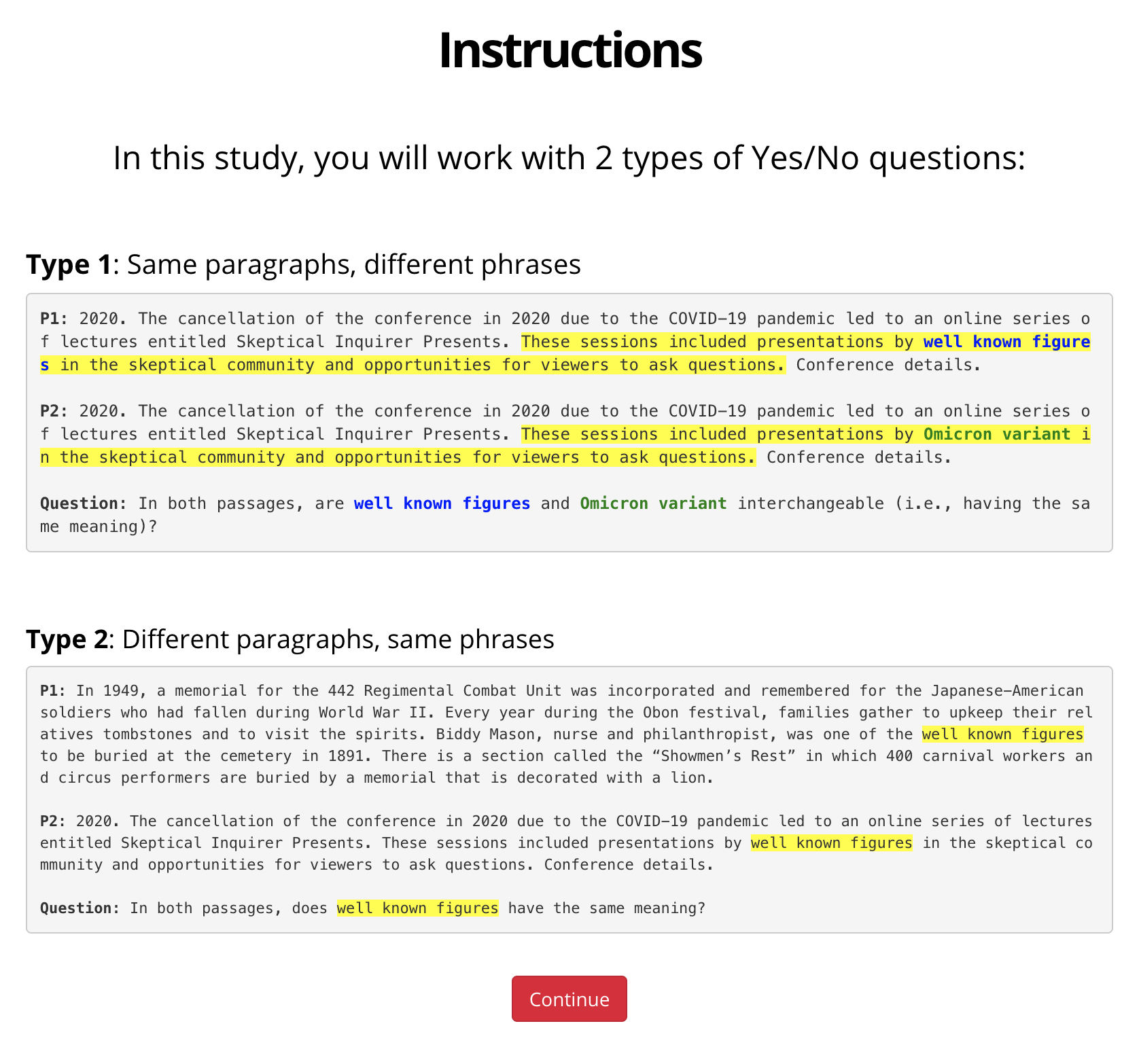}}  
  \caption{Detailed instructions given to MTurkers}
\end{subfigure}
\begin{subfigure}{.5\textwidth}
  \centering
  \mybox{YellowGreen}{\includegraphics[width=.95\linewidth]{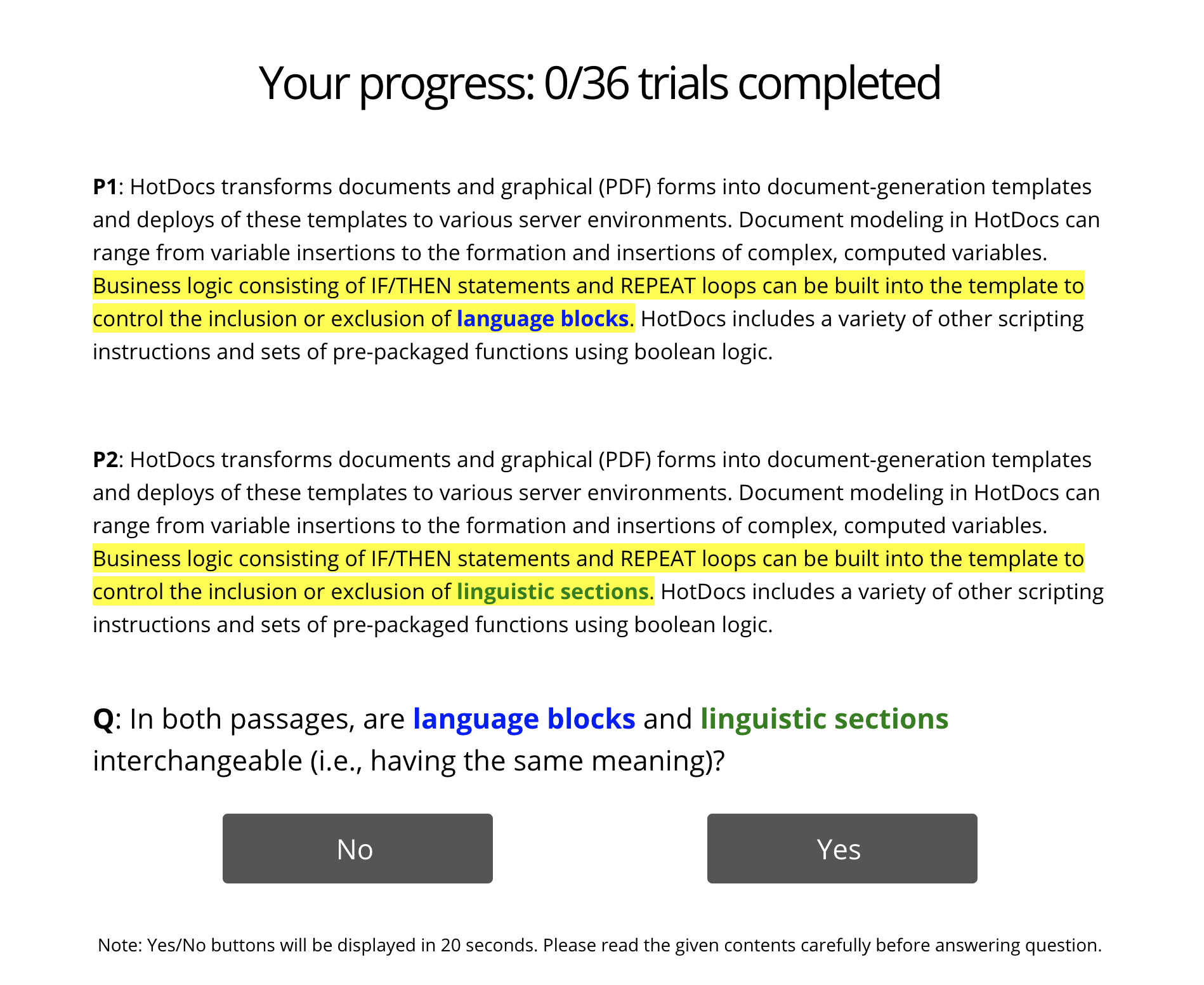}}  
  \caption{Upon completion of training stage, MTurkers need to correctly answer the first 5 out of 6 questions to be invited to verify annotations from Upwork experts.}
\end{subfigure}
\newline
\begin{subfigure}{.5\textwidth}
  \centering
  \mybox{YellowGreen}{\includegraphics[width=.95\linewidth]{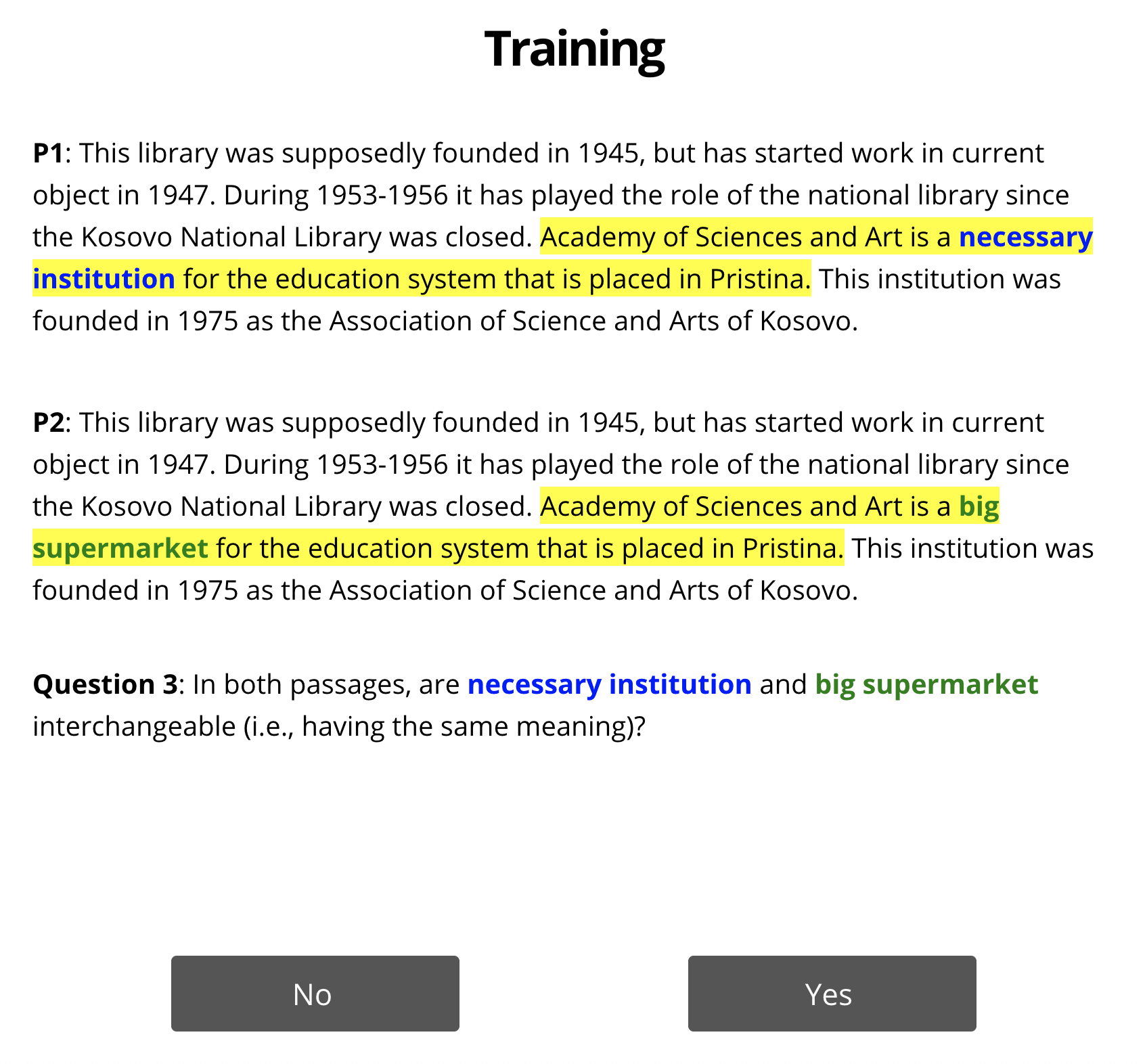}}
  \caption{Verification of paraphrases via type-1 question.}
\end{subfigure}
\begin{subfigure}{.5\textwidth}
  \centering
  \mybox{YellowGreen}{\includegraphics[width=.95\linewidth]{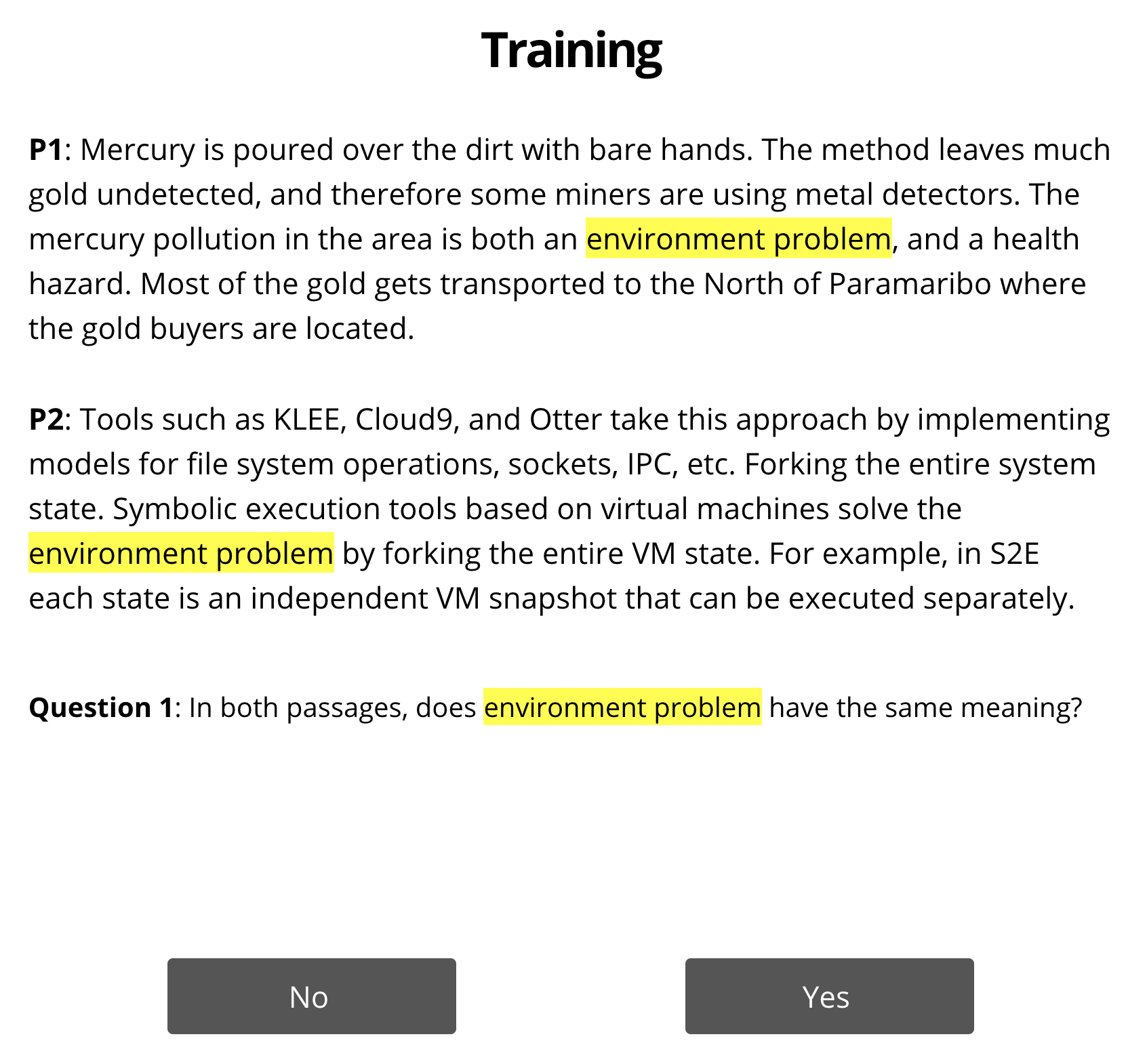}}
  \caption{Verification of Yes/No labels via type-2 question.}
\end{subfigure}
\newline
\begin{subfigure}{.5\textwidth}
  \centering
  \mybox{YellowGreen}{\includegraphics[width=.95\linewidth]{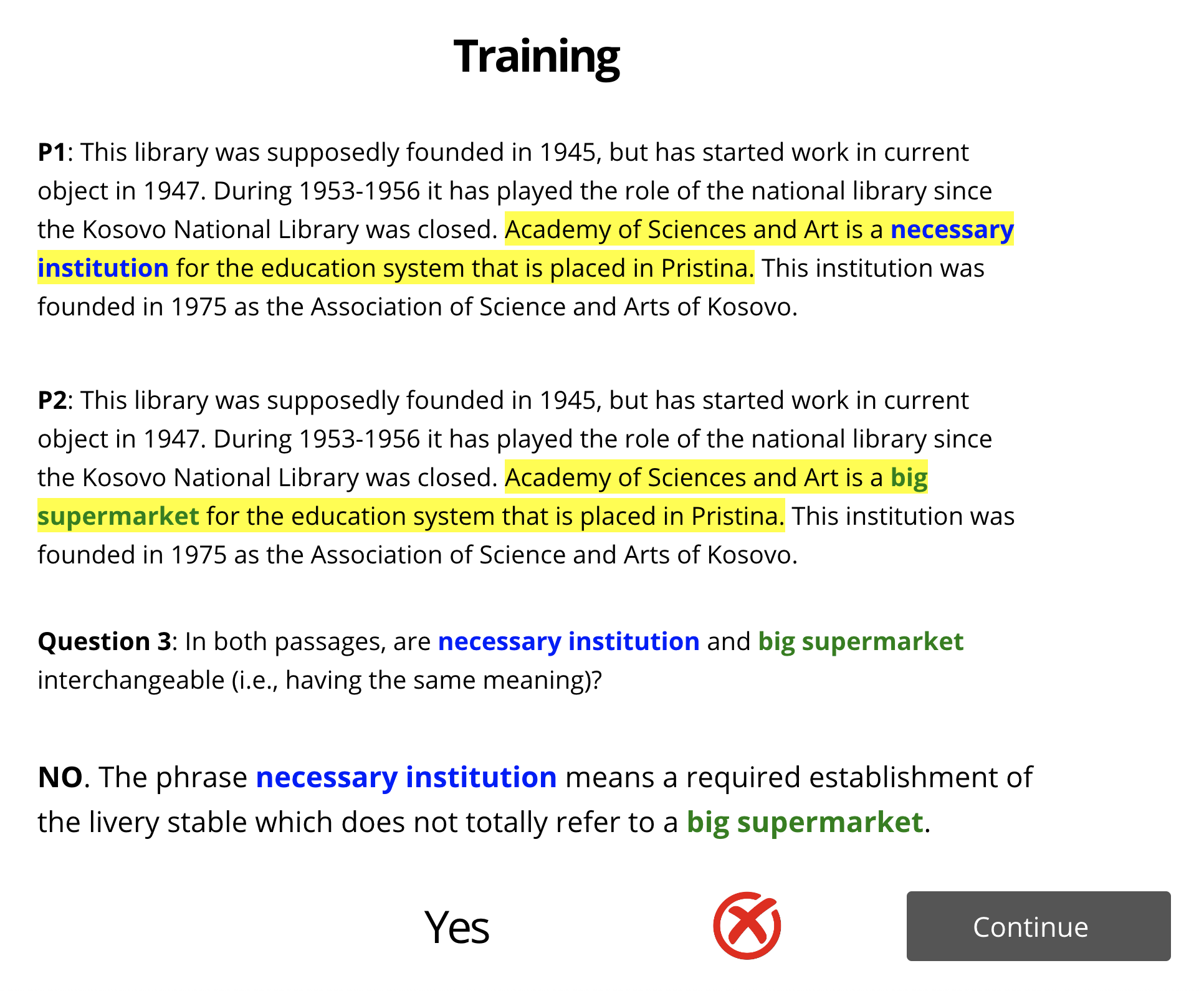}}  
  \caption{Feedback is given when MTurkers give a wrong answer.}
\end{subfigure}
\begin{subfigure}{.5\textwidth}
  \centering
  \mybox{YellowGreen}{\includegraphics[width=.95\linewidth]{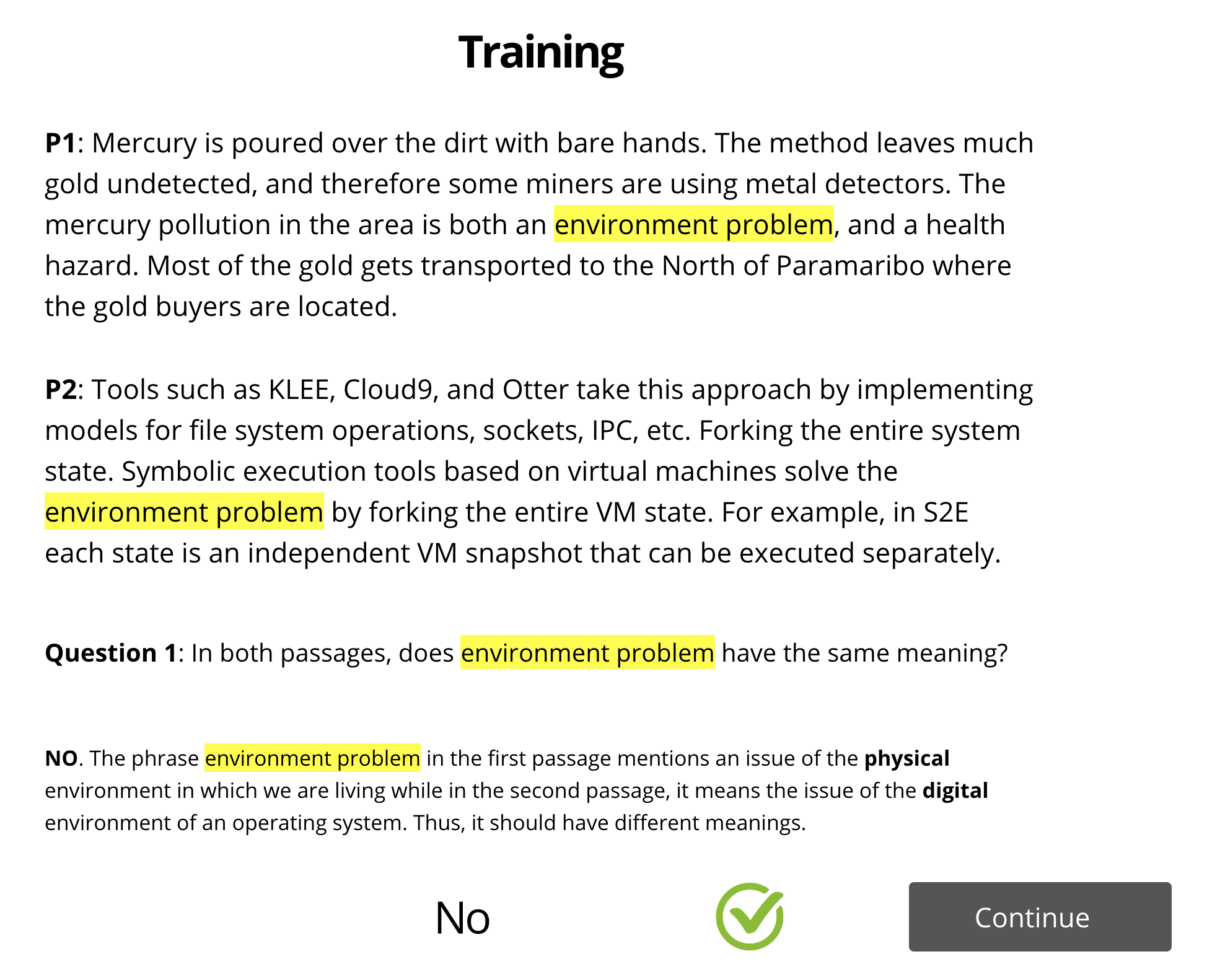}}  
  \caption{or even a right answer.}
\end{subfigure}
\caption{Gorilla layouts shown to MTurkers to verify annotations in the first round.}
\label{fig:appendix_gorilla_layout}
\end{figure*}

Participants are given detailed instructions along with 5 practice samples to get familiar with the task (Fig.~\ref{fig:appendix_gorilla_layout}).
They need to pass an evaluation checkpoint including 6 questions randomly sampled from our verified question bank in order to start working with sets of 30 questions.
With this approach, all examples in the dataset are verified once and as a result, 22,496/30,042 queries ($\sim$74.88\%) and 10,043/15,021 Yes/No answers ($\sim$66.86\%) accepted by MTurkers are considered high quality since they are annotated by a writing expert and confirmed by a qualified English native speaker.
The remaining 7,546 queries and 4,978 Yes/No answers rejected that are passed to another group of 5 writing experts for confirmation.

\subsection{Round 2: Verification by Upwork experts}
\label{sec:expert_evaluation}

We hired another set of 5 writing experts from Upwork (Upwork verifiers) with an hourly rate of \$25-40/hour to verify 12,524 examples rejected by MTurk verifiers, \ie, at an average cost of approximately \$0.26 per example.
See a sample assignment given to an Upwork expert in 
\cite{upwork_verification_samples}.

We rely on IAA to decide whether to accept or reject an example.
Specifically, we use the \emph{same question types} as shown to MTurk verifiers in the previous step and see whether these Upwork verifiers agree with the Upwork annotators to keep this example or with MTurk verifiers to reject it.
We find that the agreement between the first- and third-round annotators are 5,829 (out of 7,546) paraphrases and 3,370 (out of 4,978) Yes/No answers in total and thus the total high-quality queries and Yes/No answers we achieve are 28,325 and 13,413, respectively.

\clearpage
\section{Data Sheet}

We follow the documentation template provided by Gebru et al. 2021 
\cite{gebru2021datasheets}.

\subsection{Motivation}

\paragraph{For what purpose was the dataset created?} Understanding phrases in context plays a vital role in solving many Natural Language Understanding (NLU) tasks such as question answering or reading comprehension. 
While there are \emph{word-sense} disambiguation datasets like WiC, no such benchmarks exist for \emph{phrases}.
Existing phrase benchmarks compare only phrases without context and some of them contain numerous phrase pairs that have lexical overlap.
The major drawback is no human annotation of how a phrase's meaning changes w.r.t the context.
This motivates us to construct a Phrase-in-Context benchmark to drive the development of contextualized phrase embeddings in NLU.

\paragraph{Who created the dataset (\eg, which team, research group) and on behalf of which entity (\eg, company, institution, organization)?} Anonymous research group.  

\subsection{Composition/collection process/preprocessing/cleaning/labeling and uses}

We describe the data construction process, annotation and verification methods in our paper (See Sec.~\ref{sec:dataset_construction} and Sec.~\ref{sec:pic}).

\subsection{Distribution}

\paragraph{Will the dataset be distributed to third parties outside the entity (\eg, company, institution, organization) on behalf of which the dataset was created?} We release three datasets PS, PR (including PR-pass and PR-page) and \psd to the public.


\paragraph{When will the dataset be distributed?} It has been released in July 2022.

\paragraph{What is the dataset format and how it can be read?} We use JSON - a widely used data format for PiC dataset and follow a scheme of HuggingFace datasets to host it.
Three datasets PS, PR and \psd in the PiC dataset are loaded as folows:

\begin{lstlisting}[language=Python]
    # The following pip command is to install the HuggingFace library "datasets": pip3 install datasets
    
    from datasets import load_dataset
    
    ps      = load_dataset("PiC/phrase_similarity")
    pr_pass = load_dataset("PiC/phrase_retrieval", "PR-pass")
    pr_page = load_dataset("PiC/phrase_retrieval", "PR-page")
    psd     = load_dataset("PiC/phrase_sense_disambiguation")
\end{lstlisting}

\paragraph{Will the dataset be distributed under a copyright or other intellectual property (IP) license, and/or under applicable terms of use (ToU)?} Our dataset is distributed under the CC-BY-NC 4.0 license.

\subsection{Maintenance}


\paragraph{Will the dataset be updated (e.g., to correct labeling errors, add new instances, delete instances)? } Yes. If we include more tasks or find any errors, we will correct the dataset. It will be updated on our website and also HuggingFace. 

\paragraph{If others want to extend/augment/build on/contribute to the dataset, is there a mechanism for them to do so?} They can contact us via email for the contribution.

\end{document}